\documentclass[journal]{IEEEtran}
\usepackage{graphicx,psfrag,verbatim,color}
\DeclareGraphicsExtensions{.jpg,.pdf,.jpeg,.png}
\usepackage[small,bf]{caption}
\usepackage{amsmath}
\usepackage{amsfonts,bm,amsthm} 
\usepackage{bbm}
\usepackage{mathtools}
\usepackage{amssymb}
\usepackage{algorithm}  
\usepackage{algorithmic} 
\usepackage{subfig}
\usepackage{stfloats}

\newcommand{\field}[1]{\mathbb{#1}}

\newcommand{\R}{\field{R}}
\newcommand{\T}{\text{T}}

\newcommand{\abs}[1]{\left\lvert#1\right\rvert}
\newcommand{\norm}[1]{\left\lVert#1\right\rVert}
\newcommand{\vect}[1]{\mathbf{#1}}

\newcommand{\Expect}{\mathop{\mathbb E{}}}

\newcommand{\st}{\text{s.t.} \quad }

\newcommand{\roundBrack}[1] { \left(#1\right)   }   
\newcommand{\squareBrack}[1]{ \left[#1\right]   }   
\newcommand{\curlyBrack}[1] { \left\{#1\right\} }   
\newcommand{\ROMAN}[1]{\uppercase\expandafter{\romannumeral#1}}

\newcommand{\CN}[1]{\mathcal{CN}\roundBrack{#1}}

\theoremstyle{definition}

\newtheorem{assumption}{Assumption}
\newtheorem{theorem}{Theorem}
\newtheorem{lemma}{Lemma}

\theoremstyle{remark}

\begin{document}

\title{Successive Convex Approximation Based Off-Policy Optimization for Constrained Reinforcement Learning}
\author{\IEEEauthorblockN{Chang~Tian\IEEEauthorrefmark{1},
		An~Liu\IEEEauthorrefmark{2},~\IEEEmembership{Senior~Member,~IEEE,}
		Guan~Huang\IEEEauthorrefmark{2},
		and~Wu~Luo\IEEEauthorrefmark{1},~\IEEEmembership{Member,~IEEE}}%
	\thanks{\IEEEauthorrefmark{1}State Key Laboratory of Advanced Optical Communication Systems and Networks, Department of Electronics, Peking University ( \{tianch, luow\}@pku.edu.cn ).}
	\thanks{\IEEEauthorrefmark{2}College of Information Science and Electronic Engineering, Zhejiang University ( \{anliu, huangguan\}@zju.edu.cn ). \textit{(Corresponding Author: An Liu)}}
}



\maketitle

\begin{abstract}
We propose a successive convex approximation based off-policy optimization (SCAOPO) algorithm to solve the general constrained reinforcement learning problem, which is formulated as a constrained Markov decision process (CMDP) in the context of average cost. The SCAOPO is based on solving a sequence of convex objective/feasibility optimization problems obtained by replacing the objective and constraint functions in the original problems with convex surrogate functions. At each iteration, the convex surrogate problem can be efficiently solved by Lagrange dual method even the policy is parameterized by a high-dimensional function. Moreover, the SCAOPO enables to reuse old experiences from previous updates, thereby significantly reducing the implementation cost when deployed in the real-world engineering systems that need to online learn the environment. In spite of the time-varying state distribution and the stochastic bias incurred by the off-policy learning, the SCAOPO with a feasible initial point can still provably converge to a Karush-Kuhn-Tucker (KKT) point of the original problem almost surely.
\end{abstract}


\section{Introduction}\label{sec: intro}
\subsection{Background}
Reinforcement learning (RL) has achieved tremendous success in many sequential decision-making problems, such as playing Go \cite{2016Go_David} and controlling highly-complex robotics systems \cite{2018Robot_OpenAI}. The agent is allowed to freely explore any behavior in a simulated environment, as long as it brings performance improvement. However, there usually exist some constraints that need to be satisfied in real-world physical systems. For example, an industrial robotic system should meet certain requirement of damage avoidance. In the future 6G wireless communications, the artificial intelligence (AI)-empowered mobile edge computing systems will realize smart resource management under some energy constraints \cite{2020Network_6G}. Naturally, the huge commercial interest of deploying the RL agent to realistic domains motivates the study of constrained reinforcement learning (CRL), also termed as safe reinforcement learning, in which the agent only enables to explore behavior in a restricted domain while maximizing the performance \cite{2015Survey_Safe_RL}. 

Constrained Markov Decision Processes (CMDP) \cite{1999CMDP} provides a principled mathematical framework of incorporating constraints for dynamic systems, and has been seen as the standardized formulation for CRL problems \cite{2019BenchmarkingSRL}. In a CMDP framework, the agent attempts to maximize its expected rewards while also ensuring constraints on expectations of auxiliary costs. To attain prominent performance in practical RL problems, modern algorithms often introduce function approximations, e.g., deep neural networks (DNN), which can effectively combat the curse of dimensionality \cite{2017PPO}. However, this incurs non-convexity on both the objective function and feasible region of the CMDP. Together with the non-stationary stochasticity of objective and constraints, it is extremely challenging to solve the CMDP with high-dimensional complex function approximation. 

There are two main types of algorithms in the RL/CRL \cite{2020P3O}: The first type is \textit{on-policy} algorithms, where the agent draws a batch of data using its current policy, have poor sample efficiency and require an extremely large number of interactions with the environment to learn an effective policy. Therefore, on-policy algorithms are restricted to apply in the simulated environment that has low simulation complexity. The second type is \textit{off-policy} algorithms, where the agent reuses data from old policies to update the current policy, are data-efficient. In many practical engineering systems, the data acquirement from environment may be expensive, e.g., the estimation of channel state information in wireless communications is quite resource-consuming. Hence, it is critical to develop off-policy RL/CRL algorithms when deploying the RL/CRL agent in online learning scenario of real-world systems, where the cost of acquiring new experiences cannot be simply ignored.

In this paper, we propose an off-policy optimization algorithmic framework, called successive convex approximation based off-policy optimization (SCAOPO), to solve the general CMDP problem in the average cost case. On the one hand, reusing old experiences from previous policies is more data-efficient. On the other hand, it also brings biased estimates of policy gradient, making it much more challenging to establish the convergence of the algorithm. Nonetheless, we managed to prove that the developed SCAOPO can still converge to a Karush-Kuhn-Tucker (KKT) point of the original CMDP problem, which indicates the SCAOPO enables to guarantee stable performance with high data efficiency. To the best of our knowledge, SCAOPO is the first general algorithmic framework with theoretical convergence guarantee for CRL under the off-policy setting.

\subsection{Related Works}
There are three major lines of research on solving the CMDP.

\textbf{Primal-dual methods:} One common approach solving the CMDP is to relax the constraints via Lagrangian multipliers and formulate an unconstrained saddle-point optimization problem. 
However, the performance of primal-dual methods heavily relies on optimally solving the unconstrained RL problem on the primal domain. Due to the non-convexity and high computation complexity, existing methods usually choose to approximately solve the primal problem at the cost of sacrificing the stable performance. 
For example, \cite{2020NPGPD} updates the primal variable once at each iteration on the direction of natural policy gradient. This leads to that the final output policy is only feasible on the average. Although \cite{2018RCPO} can guarantee the convergence to a local saddle-point under some stability assumptions, \cite{2018RCPO} is on-policy and the primal problem needs to be solved to a stationary point. This indicates that \cite{2018RCPO} has to solve an entire RL problem at each iteration, which incurs huge computation complexity. \cite{2017Risk-sensitive} presents a risk-sensitive algorithm but suffers from similar drawbacks as \cite{2018RCPO}. 
To mitigate the oscillations during training or promote the statistical efficiency, \cite{2020Lagrangian_PID} and \cite{2020Optimistic_PrimalDual_PPO} propose a proportionally controlled Lagrangian method and an optimistic primal-dual proximal policy optimization algorithm, respectively. But these two works have no theoretical convergence guarantee.

\textbf{Constrained Policy Optimization (CPO):} CPO \cite{2017CPO} updates the policy within a trust-region framework. It can guarantee the performance improvement and near-constraint satisfaction at each iteration based on the performance difference bound. To alleviate the high complexity of computing the inverse of Fisher information matrix and the recovery step at each iteration, \cite{2020First-Order-CPO} simplifies the update of CPO iteration by merely using the first-order information. A projection-based CPO algorithm is proposed in \cite{2020Projection-CPO} in order to improve the convergence properties of CPO. However, all the CPO-based methods cannot theoretically assure a strictly feasible result.  

\textbf{Other methods:} \cite{2019Lyapunov} and \cite{2020Backward_Value} propose methods solving CMDP based on the Lyapunov functions and backward value functions, respectively. They both convert the trajectory-level constraints to localized state-based constraints, thus often induce high computation complexity in the continuous CRL problems. \cite{2018Safety_Layer} directly adds to the policy a safety layer thereby heuristically satisfying the constraints. All the above approaches have no theoretical guarantee for the feasibility. Recently, \cite{2019_NIPS_ConvergentPolicyOptimization} has proposed a policy optimization method for the CRL problems following from our previous work \cite{2019CSSCA}. However, this method is not applicable to the off-policy setting.

\subsection{Contributions}
All of above existing works have assumed episodic tasks with on-policy sampling, in which the system can be reinitialized at any time and the agent enables to generate arbitrarily long trajectories of experiences to compute estimates at each iteration. However, real-world engineering systems/environments usually cannot be manipulated at will, and the data sampling is not free. Moreover, existing methods except for \cite{2019_NIPS_ConvergentPolicyOptimization} do not have theoretical guarantee for the convergence to KKT points of the general CMDP, or even the convergence to a feasible point. To overcome above restrictions, we propose a novel SCAOPO algorithmic framework in this paper. The main contributions are summarized below.
\begin{itemize}
	\item \textbf{General off-policy optimization algorithmic framework:} By reusing the old experiences, SCAOPO only requires to sample one (or several) new data at each iteration, which significantly reduces the implementation cost in realistic systems and improves the data efficiency. At each update of SCAOPO, we merely needs to solve a convex objective/feasibility optimization problem obtained by replacing the objective and constraint functions in the original problems with convex surrogate functions. Moreover, the convex surrogate problem can be efficiently solved in parallel by using Lagrange-dual methods. 
	\item \textbf{Convergence proof:} Although the off-policy sampling leads to the estimation bias at each policy iteration, the proposed SCAOPO can still provably converge to a KKT point of the original CMDP problem when the initial point is feasible. Specifically, based on the analysis of stochastic bias, we first prove the asymptotic consistency, i.e., the values and gradients of surrogate functions asymptotically match the original objective/constraint functions at the current iteration. Then we show that all limiting points must be feasible with probability 1 (w.p.1), and every limiting point must be a KKT point of the convex subproblem. Together with the asymptotic consistency of surrogate functions, it can be shown that any limiting point of the SCAOPO algorithm is also a KKT point of the original problem w.p.1.
\end{itemize}

The rest of the paper is organized as follows. The problem formulation is given in Section \ref{sec: problem formulation}, together with some application examples. The SCAOPO algorithmic framework and the convergence analysis are presented in Sections \ref{sec: algo framework} and \ref{sec: convergence analysis}, respectively. Section \ref{sec: applications} provides simulation results. Finally, we concluded the paper in Section \ref{sec: conclusion}.

\section{Problem Formulation}\label{sec: problem formulation}
We first introduce some preliminaries of Markov decision process (MDP). A MDP can be represented as a tuple $ \big(\mathcal{S}, \mathcal{A}, R, P\big) $, where $ \mathcal{S} $ is the state space, $ \mathcal{A} $ is the action space, $ R: \mathcal{S}\times\mathcal{A}\to\mathbb{R} $ is the reward function, and $ P: \mathcal{S}\times\mathcal{A}\times\mathcal{S}\to\squareBrack{0,1} $ is the transition probability function with $ P\roundBrack{s^{'}|s, a} $ denoting transitioning to state $ s^{'} $ from the previous state $ s $ given that the agent took action $ a $ in $ s $. A policy $ \pi:\mathcal{S}\to\mathbf{P}\roundBrack{\mathcal{A}} $ is a map from states to probability distributions over actions, with $ \pi\roundBrack{a|s} $ denoting the probability of selecting action $ a $ in state $ s $. Because of the curse of dimensionality, modern RL algorithms, e.g., deep reinforcement learning (DRL)-based algorithms, usually parameterize the policy by function approximations with high representation capability, e.g., DNN. The parameterized policy with function approximation $ \theta $ is denoted as $ \pi_{\theta} $ in this paper.

CMDP is considered as a standardized formulation modeling CRL problems. CMDP is an MDP with additional constraints that restrict the set of allowable policies. In particular, the MDP is augmented with a set of auxiliary cost functions $ C_1, \dots, C_m $ with each one $ C_i: \mathcal{S}\times\mathcal{A}\to\mathbb{R} $ mapping state-action pairs to costs. The goal of general CRL problems is to solve the following CMDP:
\begin{subequations}\label{original_problem}
\begin{align}
\min_{\theta\in\Theta}\quad & J_0\roundBrack{\theta}\triangleq\lim_{L\to\infty}\frac{1}{L} \Expect_{p_s\sim\pi_{\theta}}\squareBrack{\sum_{l=0}^{L-1} C_0\roundBrack{s_l, a_l} }\\
\st & J_i\roundBrack{\theta}\triangleq\lim_{L\to\infty}\frac{1}{L} \Expect_{p_s\sim\pi_{\theta}}\squareBrack{\sum_{l=0}^{L-1} C_i\roundBrack{s_l, a_l} } - c_i\leq 0, \forall i,
\end{align}
\end{subequations}
where we use $ C_0\roundBrack{s_l, a_l} = -R\roundBrack{s_l, a_l} $ for the conciseness of elaboration, $ \theta\in\Theta $ is the policy parameter with $ \Theta $ being the parameter space,  $ p_s $ denotes a trajectory $ p_s = \curlyBrack{s_0, a_0, s_1, \dots} $ and $ p_s\sim\pi_{\theta} $ indicates the distribution over trajectories depends on the policy $ \pi_{\theta} $, i.e., $ a_l\sim\pi_{\theta}\roundBrack{\cdot|s_l}, s_{l+1}\sim P\roundBrack{\cdot|s_l, a_l} $, and $ c_1, \dots, c_m $ denote the limits. We make the following assumptions on the problem structure.
\begin{assumption}\label{Problem Assumptions}
	 \textit{(Assumptions of Problem \eqref{original_problem}):}
	 \begin{enumerate}
	 	\item $ \Theta\subseteq\mathbb{R}^{n_{\theta}} $ for some positive integer $ n_{\theta} $. Moreover, $ \Theta $ is compact and convex.
	 	\item State space $ \mathcal{S}\subseteq\mathbb{R}^{n_s} $ and action space $ \mathcal{A}\subseteq\mathbb{R}^{n_a} $ are both compact sets for some positive integers $ n_s $ and $ n_a $.
	 	\item For any $ i\in\curlyBrack{0,\dots,m} $, $ C_i $, the derivative and the second-order derivative of $ J_i\roundBrack{\theta} $ are uniformly bounded. Moreover, the policy $ \pi_{\theta} $ is Lipschitz continuous over the parameter $ \theta $.
	 	\item There exist constants $ m_c > 0 $ and $ \rho\in\roundBrack{0,1} $ such that 
	 	\begin{equation*}
	 	\sup_{s\in\mathcal{S}}\norm{\mathbf{P}\roundBrack{S_t|S_0=s} - \mathbf{P}_{\pi_{\theta}}}_{TV}\leq m_c\rho^t, \forall t\geq 0,	
	 	\end{equation*}
	 	\footnote{We denote $ S_t $ and $ A_t $ as the random variables of state and action at time $ t $. $ \mathbf{P}\roundBrack{S_t} $ is the state distribution at time $ t $.}
	 	where $ \mathbf{P}_{\pi_{\theta}} $ is the stationary state distribution under policy $ \pi_{\theta} $ and $ \norm{\mu - v}_{TV} = \int_{s\in\mathcal{S}}\abs{\mu\roundBrack{\mathrm{d}s} - v\roundBrack{\mathrm{d}s}} $ denotes the total-variation distance between the probability measures $ \mu $ and $ v $.
	 \end{enumerate}	
\end{assumption}

The first assumption is trivial in CRL problems. Assumption \ref{Problem Assumptions}-2) lets Problem \eqref{original_problem} consider a more general scenario, in which the state and action spaces can be continuous. Assumption \ref{Problem Assumptions}-3) indicates the Lipschitz continuity of $ J_i\roundBrack{\theta} $ and $ \nabla J_i\roundBrack{\theta}, \forall i\in\curlyBrack{0,\dots,m} $ over parameter $ \theta $, which is usually assumed in the rigorous convergence analysis of RL algorithms \cite{2002Convergent_API,2019Finite_Sample_Analysis_SARSA}. The policy Lipschitz continuity over the policy parameter is also standard as in \cite{2002Convergent_API,2019Finite_Sample_Analysis_SARSA} and holds for a number of function approximations, e.g., fully-connected DNN with \textit{Tanh} activation. Assumption \ref{Problem Assumptions}-4) indicates that the stationary state distribution under policy $ \pi_{\theta} $ exists and is independent of the initial state, which is a standard ergodicity assumption when considering problems without episode boundaries \cite{2019Finite_Sample_Analysis_SARSA}\cite[Chapter 13.6]{2018Sutton_RL_book}.  Problem \eqref{original_problem} embraces many important applications, which are specifically elaborated in the following.

\textit{Example 1 (Delay-Constrained Power Control for Downlink Multi-User MIMO (MU-MIMO) system):} Consider a downlink MU-MIMO system with one base station (BS) serving $K$ single-antenna users. The BS is equipped with $N_{t}\geq K$ transmit antennas, and it maintains $K$ data queues for the bursty traffic flows to each user. Suppose that time is divided into time slots of duration $\tau$ indexed by $t$. The queue dynamic of the $ k $-th user $Q_{k}\left(t\right), \forall k$ is given by
\[
Q_{k}\left(t\right)=\left[Q_{k}\left(t-1\right)+A_{k}\left(t\right)\tau-R_{k}\left(t\right)\tau\right]^{+},
\]
where $\left[x\right]^{+}\triangleq\max\left\{ 0,x\right\} $, $A_{k}$ represents the rate of random arrived data with $\mathbb{E}\left[A_{k}\right]=\lambda_{k}$ and $R_{k}$ denotes the transmitted data rate that is given by  
\[
R_{k}=B\log_{2}\left(1+\frac{p_k\left|\mathbf{h}^{\text{H}}_{k}\mathbf{v}_{k}\right|^{2}}{\sum_{j\neq k}p_j\left|\mathbf{h}^{\text{H}}_{k}\mathbf{v}_{j}\right|^{2}+\sigma_{k}^{2}}\right),\forall k,
\] 
where we omit the time slot index for conciseness, $ B $ is the bandwidth, $ \vect{h}_k $ is the downlink channel state information (CSI) of user $ k $, $\sigma_{k}^{2}$ is the noise power of user $k$ and $ p_k$ denotes the allocated power to user $ k $, $ \vect{v}_k $ is the $ k $-th column of  $\mathbf{V}=\left[\mathbf{v}_{1},...,\mathbf{v}_{K}\right]\in\mathbb{C}^{N_{t}\times K}$, which is the normalized regularized zero forcing (RZF) precoder \cite{2005RZF}, i.e., $ \mathbf{V} = \mathbf{H}^{\text{H}}\left(\mathbf{H}\mathbf{H}^{\text{H}} + \alpha_{Z}\mathbf{I}\right)^{-1}\mathbf{\Lambda}^{1/2} $, where $ \alpha_{Z} $ is the regularization factor, $\mathbf{H}=\left[\mathbf{h}_{1},\ldots,\mathbf{h}_{K}\right]^{\text{H}}\in\mathbb{C}^{K\times N_{t}}$, $ \mathbf{\Lambda}^{1/2} = \text{Diag}\left(\left[ \left\lVert\bar{\mathbf{v}}_1\right\rVert^{-1},\dots,\left\lVert\bar{\mathbf{v}}_K\right\rVert^{-1}  \right]\right) $ and $ \bar{\mathbf{v}}_k $ is the $ k $-th column of $ \bar{\mathbf{V}}\triangleq \mathbf{H}^{\text{H}}\left(\mathbf{H}\mathbf{H}^{\text{H}} + \alpha_{Z}\mathbf{I}\right)^{-1} $.

The objective is to obtain a power allocation and regularization factor control policy that can minimize the long-term average power consumption as well as satisfying the long-term delay constraint for each user. In particular, the BS at the $ t $-th time slot obtains the state information  $s_t=\left\{\mathbf{Q}\left(t\right),\mathbf{H}\left(t\right)\right\} $, where $\mathbf{Q}=\left[Q_{1},\ldots,Q_{K}\right]^{\text{T}}\in\R^{K}$ is the queue state information (QSI). Then the action at the $ t $-th time slot $a_t = \left\{\mathbf{p}\left(t\right), \alpha_{Z}\left(t\right)\right\}$, where $ \mathbf{p}\triangleq\left[p_1,\dots,p_K\right]^{\text{T}}\in\R^{K} $, is obtained according to the policy $ \pi_{\theta}\left(s_t\right) $. The problem of designing power control policy for the downlink MU-MIMO can be formulated as 
\begin{subequations}\label{prob: power control}
	\begin{align}
	\min_{\theta\in\Theta}\quad & J_0\roundBrack{\theta}\triangleq\lim_{T\to\infty}\frac{1}{T} \Expect\squareBrack{\sum_{t=1}^{T}\sum_{k=1}^{K}p_k\roundBrack{t} }\\
	\st & J_k\roundBrack{\theta}\triangleq\lim_{T\to\infty}\frac{1}{T} \Expect\squareBrack{\sum_{t=1}^{T} \frac{Q_k\roundBrack{t}}{\lambda_{k}} } - c_k\leq 0,\; \forall k,
	\end{align}
\end{subequations}
where $c_{1},\ldots,c_{K}$ denote the maximum allowable average delay for each user.

\textit{Example 2 (Constrained Linear-Quadratic Regulator):} Linear-quadratic regulator (LQR) is one of the most fundamental problems in control theory and has attracted great attention in the RL community \cite{2019LQR_RL}. In the LQR setting, the state dynamic is linear and the cost function is quadratic. Specifically, denote $ s_t\in\mathbb{R}^{n_s} $ and $ a_t\in\mathbb{R}^{n_a} $ as the state and action at the time $ t $, respectively. The next state $ s_{t+1} $ and the cost $ C_0\roundBrack{s_t, a_t} $ are given by
\begin{align*}
	s_{t+1} &= As_t + Ba_t + \epsilon_t, \\ 
	C_0\roundBrack{s_t, a_t} &= s_t^{\text{T}} Q_0 s_t + a_t^{\text{T}} R_0 a_t, 
\end{align*}
where $ A\in\mathbb{R}^{n_s\times n_s} $ and $ B\in\mathbb{R}^{n_s\times n_a} $ are transition matrices, $ \curlyBrack{\epsilon_t} $ is the process noise and $ Q_0\in\mathbb{R}^{n_s\times n_s} $ and $ R_0\in\mathbb{R}^{n_a\times n_a} $ are cost matrices. Then the aim of constrained LQR is to find a policy $ \pi_{\theta} $ that satisfies:
\begin{subequations}\label{prob: CLQR}
	\begin{align}
	\min_{\theta\in\Theta}\quad & J_0\roundBrack{\theta}\triangleq\lim_{T\to\infty}\frac{1}{T} \Expect\squareBrack{\sum_{t=0}^{T-1} C_0\roundBrack{s_t, a_t} }\\ 
	\st & J_i\roundBrack{\theta}\triangleq\lim_{T\to\infty}\frac{1}{T} \Expect\squareBrack{\sum_{t=0}^{T-1} C_i\roundBrack{s_t, a_t} } - c_i\leq 0, \forall i, 
	\end{align}
\end{subequations}
where $ i $ indicates the cost matrices $ Q_i $ and $ R_i $.

\section{SCAOPO Algorithmic Framework}\label{sec: algo framework}
\subsection{Challenges of Solving Problem \eqref{original_problem}}
Since Problem \eqref{original_problem} is a general non-convex problem, we focus on designing an efficient algorithm to find a KKT point of Problem \eqref{original_problem}. However, Problem \eqref{original_problem} is a CMDP problem where the state distribution is time-varying because of the updates of policy. Together with the stochastic nature of objective and constraints, it is hard to ensure any limiting point of the algorithm is a KKT point almost surely. 

The state distribution and the environment model (i.e., the transition probability function) are usually not known a priori. Therefore, the estimates have to be obtained from the measurements by the agent interacting with the environment, which requires to design a low-cost estimation strategy. Moreover, we prefer to devise a low-complexity algorithm at each policy update, such that the policy parameters can be efficiently updated even using a high-dimensional complex function approximation, such as DNN.

\subsection{Summary of SCAOPO Algorithm}
We propose a successive convex approximation based off-policy optimization (SCAOPO) algorithm to solve Problem \eqref{original_problem}. Specifically, at the $ t $-th iteration, the policy parameter $ \theta $ is updated by solving a convex optimization problem obtained by replacing the objective and constraint functions $ J_i\roundBrack{\theta}, i = 0,\dots,m $, with their convex surrogate functions $ \bar{J}_i^t\roundBrack{\theta}, i = 0,\dots,m $. The surrogate function $ \bar{J}_i^t\roundBrack{\theta} $ can be seen as a convex approximation of $ J_i\roundBrack{\theta} $ based on the $ t $-th iterate $ \theta^t $, which has the following form:
\begin{equation}\label{eqn: surrogate function}
	\bar{J}_i^t\roundBrack{\theta} = \hat{J}_i^t + \roundBrack{\hat{g}_i^t}^{\T}\roundBrack{\theta - \theta^t} + \varsigma_i\norm{\theta - \theta^t}_2^2, \forall i,
\end{equation}  
where $ \varsigma_i $ is a positive constant and $ \hat{J}_i^t\in\mathbb{R} $ and $ \hat{g}_i^t\in\mathbb{R}^{n_{\theta}} $ are the estimate of function value $ J_i\roundBrack{\theta} $ and the estimate of gradient $ \nabla J_i\roundBrack{\theta} $ at the $ t $-th iteration, which are updated by
\begin{subequations}
\begin{align}
	\hat{J}_i^t &= \roundBrack{1-\alpha_t}\hat{J}_i^{t-1} + \alpha_t\tilde{J}_i^t, \\
	\hat{g}_i^t &= \roundBrack{1-\alpha_t}\hat{g}_i^{t-1} + \alpha_t\tilde{g}_i^t, \label{eqn: gradient_moving_avg}
\end{align}
\end{subequations}
where $ \curlyBrack{\alpha_t} $ is a decreasing sequence satisfying the Assumption \ref{Assmp: stepsizes} in Section \ref{sec: convergence analysis} and $ \tilde{J}_i^t $ and $ \tilde{g}_i^t $ are the new realizations of estimates of function value and gradient at the $ t $-th iteration, whose specific forms are given in Section \ref{sec: estimation strategy}.  

After constructing the surrogate functions $ \bar{J}_i^t\roundBrack{\theta}, i = 0,\dots,m $, the optimal solution $ \bar{\theta}^t $ of the following problem is solved:
\begin{align}\label{prob: objective update}
	\bar{\theta}^t = \arg\min_{\theta\in\Theta}\quad & \bar{J}_0^t\roundBrack{\theta} \notag\\
	\st & \bar{J}_i^t\roundBrack{\theta}\leq 0,\; i=1,\dots,m.
\end{align}
 Problem \eqref{prob: objective update} can be viewed as a convex approximation of Problem \eqref{original_problem}. Note that Problem \eqref{prob: objective update} is not necessarily feasible. If Problem \eqref{prob: objective update} turns out to be infeasible, the optimal solution $ \bar{\theta}^t $ of the following convex problem is solved:
\begin{align}\label{prob: feasible update}
\bar{\theta}^t = \arg\min_{\theta\in\Theta, x}\quad & x \notag\\
\st & \bar{J}_i^t\roundBrack{\theta}\leq x,\; i=1,\dots,m,
\end{align}
which minimizes the constraint violations. Given $ \bar{\theta}^t $ in one of the above two cases (in which \eqref{prob: objective update} and \eqref{prob: feasible update} are called objective update and feasible update, respectively), $ \theta $ is updated according to
\begin{equation}\label{eqn: policy update}
	\theta^{t+1} = \roundBrack{1-\beta_t}\theta^t + \beta_t\bar{\theta}^t,
\end{equation}
where $ \curlyBrack{\beta_t} $ is a decreasing sequence satisfying the Assumption \ref{Assmp: stepsizes} in Section \ref{sec: convergence analysis}. The overall algorithm is summarized in Algorithm 1.

\subsection{Estimation of $ \tilde{J}_i^t $ and $ \tilde{g}_i^t $}\label{sec: estimation strategy}
We propose an off-policy estimation strategy in which $ \tilde{J}_i^t $ and $ \tilde{g}_i^t $ can be obtained by reusing old experiences, such that the SCAOPO algorithm is well-suited to the online learning scenario when deployed in real-world systems. In particular, the agent stores the latest $ 2T $ experiences, i.e., $ \varepsilon_t = \Big\{s_{t-2T+1}, a_{t-2T+1}, \curlyBrack{C^{'}_i\roundBrack{s_{t-2T+1}, a_{t-2T+1}}}_{i = 0,\dots,m}, \\
s_{t-2T+2}, \dots, s_t, a_t, \curlyBrack{C^{'}_i\roundBrack{s_t,a_t}}_{i = 0,\dots,m}\Big\} $, where $ C^{'}_0\roundBrack{s_t,a_t} = C_0\roundBrack{s_t,a_t} $ and $ C^{'}_i\roundBrack{s_t,a_t} = C_i\roundBrack{s_t,a_t} - c_i, i = 1,\dots,m $. At the $ t $-th iteration, the agent obtains the new state $ s_t $, generates an action $ a_t $ by the policy $ \pi_{\theta^t} $, applies it to the environment and the environment feedbacks costs $ \curlyBrack{C^{'}_i\roundBrack{s_t, a_t}}_{i = 0,\dots,m} $. Then the agent updates the storage of experiences.

The reason we save $ 2T $ number of experiences is the estimation of gradient $ \tilde{g}_i^t, \forall i $ involves the estimation of Q-value, which needs to be obtained from a trajectory of experiences. Based on the accumulated experiences $ \varepsilon_t $, the new realization of estimate of function value at the $ t $-th iteration $ \tilde{J}_i^t, \forall i $ is given by the sample average method as
\begin{equation}\label{eqn: function value estimate}
	\tilde{J}_i^t = \frac{1}{2T}\sum_{l=1}^{2T}C^{'}_i\roundBrack{s_{t-2T+l}, a_{t-2T+l}}, i=0,\dots,m.
\end{equation}

According to the policy gradient theorem \cite{2018Sutton_RL_book,1999Policy_Gradient}, the gradient of $ J_i\roundBrack{\theta}, \forall i $ is
\begin{equation}\label{eqn: policy gradient}
\nabla J_i\roundBrack{\theta} = \Expect_{\substack{ s\sim\mathbf{P}_{\pi_{\theta}}\\ a\sim\pi_{\theta}\roundBrack{\cdot|s} }}\squareBrack{Q_i^{\pi_{\theta}}\roundBrack{s,a}\nabla\log\pi_{\theta}\roundBrack{a|s}},
\end{equation}
where 
\begin{equation}\label{eqn: Q-value}
Q_i^{\pi_{\theta}}\roundBrack{s,a} = \Expect_{p_s\sim\pi_{\theta}}\squareBrack{\left.\sum_{l=0}^{\infty} \roundBrack{C^{'}_i\roundBrack{s_l, a_l} - J_i\roundBrack{\theta}} \right| S_0 = s, A_0 = a }
\end{equation}
is the Q-value function under policy $ \pi_{\theta} $. We adopt the idea of sample average. Then the new realization of estimate of gradient at the $ t $-th iteration $ \tilde{g}_i^t, \forall i $ is given by
\begin{align}\label{eqn: gradient estimate}
\tilde{g}_i^t = \frac{1}{T}\sum_{l=1}^{T}&\hat{Q}_i^{t-2T+l}\roundBrack{s_{t-2T+l}, a_{t-2T+l}}\notag\\ &\nabla\log\pi_{\theta^t}\roundBrack{a_{t-2T+l}|s_{t-2T+l}},
\end{align}
where 
\begin{align}
\hat{Q}_i^{t-2T+l}&\roundBrack{s_{t-2T+l}, a_{t-2T+l}} = \notag\\
&\sum_{l^{'} = 0}^{T-1}\roundBrack{ C^{'}_i\roundBrack{s_{t-2T+l+l^{'}}, a_{t-2T+l+l^{'}}} - \hat{J}_i^t } \notag
\end{align}
is the estimate of Q-value starting from state $ s_{t-2T+l} $ and action $ a_{t-2T+l} $, which is obtained by using a trajectory of $ T $ experiences starting from state $ s_{t-2T+l} $ and action $ a_{t-2T+l} $.

Note that we can also generate multiple new experiences at each iteration in order to accelerate the convergence of SCAOPO algorithm. The number of newly added data at each iteration is referred to as batch size in this paper. It can be seen from the above estimation procedure that the estimates of function value and gradient at each iteration only require one (or several) new data samples and some simple operations of sample average. This estimation strategy is data-efficient and low-cost.
 Moreover, we do not need to know the specific form of cost functions $ C_i, \forall i $, but only need to have access to the outcomes of cost functions given the state-action pairs. This makes our algorithmic framework is applicable to scenarios where the cost functions $ C_i, \forall i $ are hard to be quantified as closed-form expressions.

\begin{algorithm}[!t] 
	\caption{Stochastic Convex Approximation Based Off-Policy Optimization Algorithm}
	\label{algo: SCAPO}
	\renewcommand{\algorithmicrequire}{\textbf{Input:}}
	\begin{algorithmic}
		\REQUIRE The two decreasing sequences $ \curlyBrack{\alpha_t} $ and $ \curlyBrack{\beta_t} $, the initial policy parameters $ \theta^0\in\Theta $ and first accumulate $ 2T $ experiences.
		\FOR{$t = 0,1,\dots$}
		\STATE Sample an action $ a_t\sim\pi_{\theta^t}\roundBrack{\cdot|s_t} $, obtain costs $ \curlyBrack{C^{'}_i\roundBrack{s_t, a_t}}_{i = 0,\dots,m} $ and update the data storage $ \varepsilon_t $.
		\STATE Estimate function values and gradients by \eqref{eqn: function value estimate} and \eqref{eqn: gradient estimate}, respectively.
		\STATE Update the surrogate functions $ \curlyBrack{\bar{J}_i^t\roundBrack{\theta}}_{i = 0,\dots,m} $ via \eqref{eqn: surrogate function}.
		\IF{Problem \eqref{prob: objective update} is feasible}
		\STATE Solve \eqref{prob: objective update} to obtain $ \bar{\theta}^t $. \textit{(Objective update)}
		\ELSE 
		\STATE Solve \eqref{prob: feasible update} to obtain $ \bar{\theta}^t $. \textit{(Feasible update)}
		\ENDIF
		\STATE Update policy parameter $ \theta^{t+1} $ according to \eqref{eqn: policy update}.
		\ENDFOR
	\end{algorithmic}
\end{algorithm}

\subsection{Efficient Solutions for Subproblems \eqref{prob: objective update} and \eqref{prob: feasible update}}
We propose a low-complexity Lagrange dual algorithm for solving the convex surrogate subproblems in \eqref{prob: objective update} and \eqref{prob: feasible update}. The reasons for using the Lagrange dual method are as follows. First, for given Lagrangian multipliers (which are also called dual variable), the problem of minimizing the Lagrangian function in the primal domain has a closed-form global minimizer. Second, the dimension of primal variable $ \theta $ is usually much larger than the dimension of dual variable, which is reflected by the number of constraints. For example, modern RL algorithms often employ DNN as the function approximation. Therefore, the optimal dual variable can be solved much more efficiently than directly solving the optimal primal variable. Since Problem \eqref{prob: objective update} and \eqref{prob: feasible update} have a similar form, we only show the solution of Problem \eqref{prob: objective update} in the following for conciseness.

The Lagrangian function for Problem \eqref{prob: objective update} is 
\begin{align}
	\mathcal{L}^t\roundBrack{\theta, \lambda} &= \bar{J}_0^t\roundBrack{\theta} + \sum_{i=1}^{m}\lambda_i\bar{J}_i^t\roundBrack{\theta} \notag\\
	&= \sum_{j=1}^{n_{\theta}} a\roundBrack{\lambda}\roundBrack{\theta_j}^2 + b_{j}\roundBrack{\lambda}\theta_j + c\roundBrack{\lambda}, \theta\in\Theta,
\end{align}
where $\lambda = \squareBrack{\lambda_1,\dots,\lambda_m}^{\T}\in\mathbb{R}_{+}^m$ are Lagrangian multipliers, $ \theta_j $ is the $ j $-th element of $ \theta $
\begin{align}
	a\roundBrack{\lambda} &= \sum_{i=0}^{m}\lambda_i\varsigma_i, \notag\\
	b_{j}\roundBrack{\lambda} &= \sum_{i=0}^{m}\lambda_i\roundBrack{ \hat{g}_{i,j}^t - 2\varsigma_i\theta_j^t },\notag\\
	c\roundBrack{\lambda} &= \sum_{i=0}^{m}\lambda_i\roundBrack{\hat{J}_i^t - \roundBrack{\hat{g}_i^t}^{\T}\theta^t + \varsigma_i\norm{\theta^t}_2^2}, \notag
\end{align}
$ \lambda_0 = 1 $, $ \theta_j^t $ and $ \hat{g}_{i,j}^t $ are the $ j $-th element of $ \theta^t $ and $ \hat{g}_{i}^t $, respectively. The dual function for Problem \eqref{prob: objective update} is
\begin{equation}\label{eqn: dual function}
	d^t\roundBrack{\lambda} = \min_{\theta\in\Theta}\ \mathcal{L}^t\roundBrack{\theta, \lambda}.
\end{equation}
The corresponding dual problem is 
\begin{equation}\label{eqn: dual problem}
\max_{\lambda\in\mathbb{R}_{+}^m}\  d^t\roundBrack{\lambda}.
\end{equation}

Thanks to the structure of convex quadratic optimization of \eqref{eqn: dual function}, the global minimizer of \eqref{eqn: dual function} for a given dual variable $ \lambda $ has a closed-form expression as
\begin{equation}
	\theta_j^{\circ}\roundBrack{\lambda} = \mathbb{P}_{\Theta_j}\squareBrack{-\frac{b_j\roundBrack{\lambda}}{2a\roundBrack{\lambda}}}, j=1,\dots,n_{\theta},
\end{equation}
where $ \mathbb{P}_{\Theta_j} $ denotes the one-dimensional projection onto the convex set $ \Theta_j $, which can be implemented in parallel. On the other hand, it can be verified that Problem \eqref{eqn: dual problem} is concave and $ \Big[\bar{J}_1^t\roundBrack{\theta^{\circ}\roundBrack{\lambda}}, \dots, \bar{J}_m^t\roundBrack{\theta^{\circ}\roundBrack{\lambda}}\Big]^{\T} $ is a subgradient of $  d^t\roundBrack{\lambda} $ at $ \lambda $. Hence, we can use the standard subgradient-based methods such as \cite{2003Sub_grad} to efficiently obtain the optimal solution $ \lambda^{\circ} $ of the dual problem in \eqref{eqn: dual problem}. Then the optimal primal solution of Problem \eqref{prob: objective update} is $ \theta^{\circ}\roundBrack{\lambda^{\circ}} $.

\section{Convergence Analysis}\label{sec: convergence analysis}
We prove that the SCAOPO algorithm presented in Algorithm \ref{algo: SCAPO} can converge to a KKT point of the Problem \eqref{original_problem}. To state the convergence result, we need to make the assumptions on the sequence of step sizes $ \curlyBrack{\alpha_t} $ and $ \curlyBrack{\beta_t} $:
\begin{assumption}\label{Assmp: stepsizes}
	\textit{(Assumptions on  $ \curlyBrack{\alpha_t} $, $ \curlyBrack{\beta_t} $):} 
	\begin{itemize}
		\item[1)] $ \alpha_t\rightarrow 0 $,  $ \frac{1}{\alpha_t}\leq O\roundBrack{t^{\kappa }} $ for some $ \kappa\in\roundBrack{0,1} $, $\sum_t\alpha_t t^{-1}<\infty$, $ \sum_t\alpha_t\roundBrack{\log^2 t \sum_{t^{'}=t-\log t}^{t}\beta_{t^{'}}}<\infty $, $ \sum_t\roundBrack{\alpha_t}^2<\infty $.
		\item[2)] $ \beta_t\rightarrow 0 $, $ \sum_t\beta_t=\infty  $, $ \sum_t\roundBrack{\beta_t}^2<\infty $
		\item[3)] $\lim_{t\rightarrow\infty}\frac{\beta_t}{\alpha_t}=0 $
	\end{itemize}
\end{assumption}
Note that the condition $ \frac{1}{\alpha_t}\leq O\roundBrack{t^{\kappa}} $ for some $ \kappa\in\roundBrack{0,1} $ is almost the same as $ \sum_t \alpha_t = \infty $, which is a common assumption in stochastic optimization algorithms \cite{2016ParallelDecomposition,2020RCSHP}. A typical choice of $ \curlyBrack{\alpha_t} $ and $ \curlyBrack{\beta_t} $ satisfying Assumption \ref{Assmp: stepsizes} is $ \alpha_t = t^{-\kappa_1} $ and $ \beta_t = t^{-\kappa_2} $, where $ \kappa_1\in\roundBrack{0.5, 1} $, $ \kappa_2\in(0.5, 1] $ and $ \kappa_1 < \kappa_2 $.

With Assumptions \ref{Problem Assumptions} and \ref{Assmp: stepsizes}, we first prove a lemma that indicates the asymptotic consistency of the function values and gradients.
\begin{lemma}\label{lemma: asymptotic consistency}
	\textit{(Asymptotic consistency of surrogate functions:)} For all $ i\in\curlyBrack{0,\dots,m} $, we have
	\begin{subequations}
		\begin{align}
		\lim_{t\to\infty}\abs{\hat{J}_i^t - J_i\roundBrack{\theta^t}} &= 0, \label{eqn: function value consistency}\\
		\lim_{t\to\infty}\norm{\hat{g}_i^t - \nabla J_i\roundBrack{\theta^t}}_2 &= 0 \label{eqn: gradient consistency}. 
		\end{align}	
	\end{subequations} 
\end{lemma}
Please refer to Appendix \ref{Proof:Lemma of consistency} for the proof. Then considering a subsequence $ \curlyBrack{\theta^{t_j}}_{j=1}^{\infty} $ converging to a limiting point $ \theta^{*} $, there exist converged surrogate functions $ \hat{J}_i\roundBrack{\theta}, i=0,\dots,m $ such that
\begin{equation}\label{eqn: converged surrogate functions}
	\lim_{j\to\infty}\bar{J}_i^{t_j}\roundBrack{\theta} = \hat{J}_i\roundBrack{\theta}, \forall\theta\in\Theta,
\end{equation}
where  
\begin{align}
\abs{\hat{J}_i\roundBrack{\theta^{*}} - J_i\roundBrack{\theta^{*}}} &= 0, \notag\\
\norm{\nabla\hat{J}_i\roundBrack{\theta^{*}} - \nabla J_i\roundBrack{\theta^{*}}}_2 &= 0 \notag. 
\end{align}	
Before the introduction of the main convergence theorem, we present the concept of Slater condition for the converged surrogate functions. 

\textbf{Slater condition for the converged surrogate functions:} Given a subsequence $ \curlyBrack{\theta^{t_j}}_{j=1}^{\infty} $ that converges to a limiting point $ \theta^{*} $, let $ \hat{J}_i\roundBrack{\theta} $ be the converged surrogate functions defined in \eqref{eqn: converged surrogate functions}. We say that the Slater condition is satisfied at $ \theta^{*} $ if there exists $ \theta\in\mbox{relint}\; \Theta $ such that
\begin{equation*}
	\hat{J}_i\roundBrack{\theta} < 0, \forall i\in\curlyBrack{1,\dots,m}.
\end{equation*}
This is a standard constraint qualification condition that is usually assumed in the proof of KKT point convergence \cite{2014SCA_PhD}.

With the Lemma \ref{lemma: asymptotic consistency} and Slater condition, we are ready to prove the main convergence result.
\begin{theorem}\label{thm: convergence of COPO}
	\textit{(Convergence of Algorithm \ref{algo: SCAPO}:)} Suppose Assumptions \ref{Problem Assumptions} and \ref{Assmp: stepsizes} are satisfied, and the initial point $\theta^{0}$ is feasible, i.e., $\max_{i\in\left\{1,...,m\right\} }J_{i}\left(\theta^{0}\right)\leq 0$. Denote $\left\{\theta^{t}\right\} _{t=1}^{\infty}$ as the iterates generated by Algorithm \ref{algo: SCAPO} with a sufficiently small initial
	step size $\beta_{0}$. Let the number of data samples $ T^t = O\roundBrack{\log t} $. Then every limiting point $\theta^{*}$ of $\left\{\theta^{t}\right\} _{t=1}^{\infty}$ satisfying the Slater condition is a KKT point of Problem \eqref{original_problem} almost surely.
\end{theorem}

Please refer to Appendix \ref{Proof:Theorem of convergence} for the proof. Note that the assumption on the sufficiently small initial step size $ \beta_0 $ is mainly for the rigorous convergence proof. Although the number of data samples used for estimation requires to approach infinity as time approaches infinity in theory, the increasing speed is on the logarithmic order, which is relatively slow. In practice, we find that using a constant number of experiences for estimation has already enabled a good convergence behavior.


Finally, we discuss the convergence behavior of Algorithm \ref{algo: SCAPO} with an infeasible initial point. In this case, it follows from the analysis in Appendix \ref{Proof:Theorem of convergence} that Algorithm \ref{algo: SCAPO} either converges to KKT points of Problem \eqref{original_problem}, or converges to the following \textit{undesired set}:
\begin{equation*}
	\overline{\Theta}_{C}^{*}=\left\{ \theta:\: J\roundBrack{\theta}>0,\:\theta\in\Theta_{C}^{*}\right\},
\end{equation*}
where $\Theta_{C}^{*}$ is the set of stationary points of the following constraint minimization problem:
\begin{equation}\label{eqn: max constraints}
\min_{\theta\in\Theta}\: J\roundBrack{\theta}\triangleq\max_{i\in\left\{ 1,...,m\right\}} J_i\roundBrack{\theta}.
\end{equation}
Thanks to the proposed feasible update \eqref{prob: feasible update}, Algorithm \ref{algo: SCAPO} may still converge to a KKT point of Problem \eqref{original_problem} even with an infeasible initial point, as long as the initial point is not close to an undesired point $\theta_{C}^{*}\in\overline{\Theta}_{C}^{*}$ such that the algorithm gets stuck in this undesired point. In practice, if we generate multiple random initial points, it is likely that in the iterates starting from these random initializers, there exist feasible points, which can be viewed as feasible initial points, thereby guaranteeing that the algorithm will converge to a KKT point of Problem \eqref{original_problem}.

\textbf{Remark 1.} \textit{(Key differences from the related works)}: Our previous work \cite{2019CSSCA} and the work \cite{2019_NIPS_ConvergentPolicyOptimization} following \cite{2019CSSCA} are closely related to the work in this paper. However, there are several key differences from the above two works:
\begin{itemize}
	\item \textbf{Different Problem Formulation:} \cite{2019CSSCA} considers the stochastic optimization problem, where the state distribution is invariant. Both \cite{2019_NIPS_ConvergentPolicyOptimization} and this paper focus on the sequential decision-making problem, where the state distribution changes along with the updates of policy (optimization variable). However, the objective and constraints in \cite{2019_NIPS_ConvergentPolicyOptimization} are formulated as the expected discounted sum of costs (called discounted cost case), while this paper takes into account the long-term average cost case. 
	\item \textbf{Different Estimation Strategy:} Both \cite{2019CSSCA} and \cite{2019_NIPS_ConvergentPolicyOptimization} employs the on-policy estimation strategy \footnote{Since the state distribution in \cite{2019CSSCA} is static, we roughly categorize it into the on-policy estimation. The definitions of on-policy and off-policy are elaborated in Section \ref{sec: intro}.}, while this paper adopts the off-policy estimation, which indicates the SCAOPO is better suited to the online learning scenario in practical engineering systems. Moreover, the definitions of function value and policy gradient of the average cost case are quite different from the discounted cost case in \cite{2019_NIPS_ConvergentPolicyOptimization}. Hence, the specific procedure of estimation need to be carefully designed.
	\item \textbf{New Challenges on Convergence Analysis:} On-policy estimation employed in \cite{2019CSSCA} and \cite{2019_NIPS_ConvergentPolicyOptimization} guarantees unbiased estimates at each iteration, thereby making the satisfaction of asymptotic consistency easily. However, SCAOPO reuses experiences from old policies, which incurs biased estimates at each iteration. Together with the time-varying state distribution, we need to introduce new techniques to analyze the convergence of function value and gradient. 
\end{itemize}

\section{Simulation Results}\label{sec: applications}
We adopt the advanced CRL algorithms, proximal policy optimization-Lagrangian (PPO-Lagrangian) \cite{2019BenchmarkingSRL} and constrained policy optimization \cite{2017CPO}, as baselines to compare with the proposed SCAOPO. For SCAOPO, we maintain the storage of $ 2T = 3000 $ experiences and the number of newly added data (batch size) at each iteration is 100 and 1000 for Example 1 and Example 2, respectively. To demonstrate the effect of reusing old experiences, we also simulate the SCAOPO with the number of newly added data (batch size) is 1000 and 3000 at each iteration, but without storing previous data samples. We employ the commonly used Gaussian policy \cite[Chapter 13.7]{2018Sutton_RL_book}, which is parameterized by a three-layer fully-connected neural network, where sizes and activation functions of two hidden layers are both 128 and \textit{Tanh} functions, respectively, while the activation function of output layer is \textit{Sigmoid} function.
 
 \begin{figure}[!t]
	\centering
	\subfloat[]{\includegraphics[width=3in]{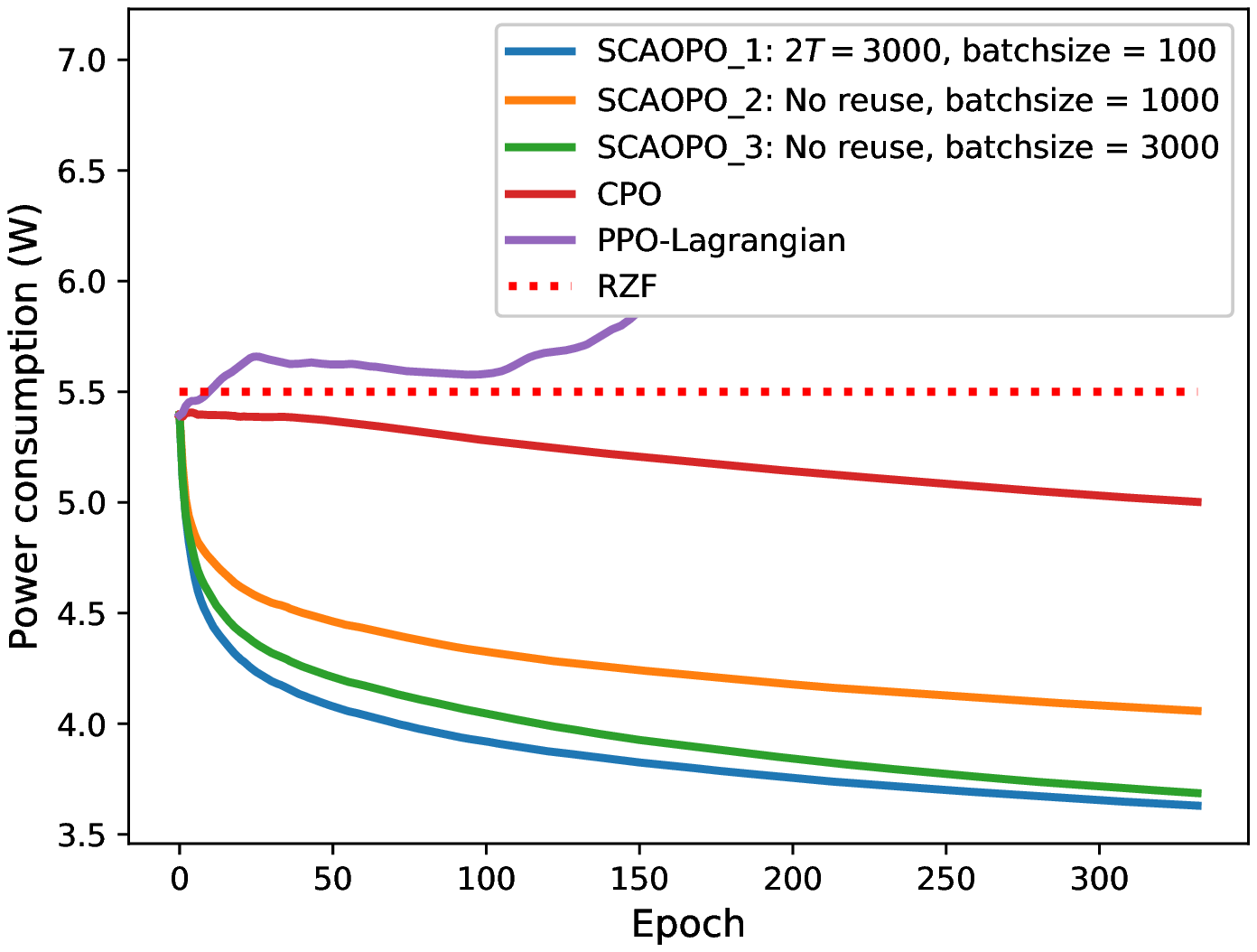}%
		\label{fig: power0} } 	\hfil
	\subfloat[]{\includegraphics[width=3in]{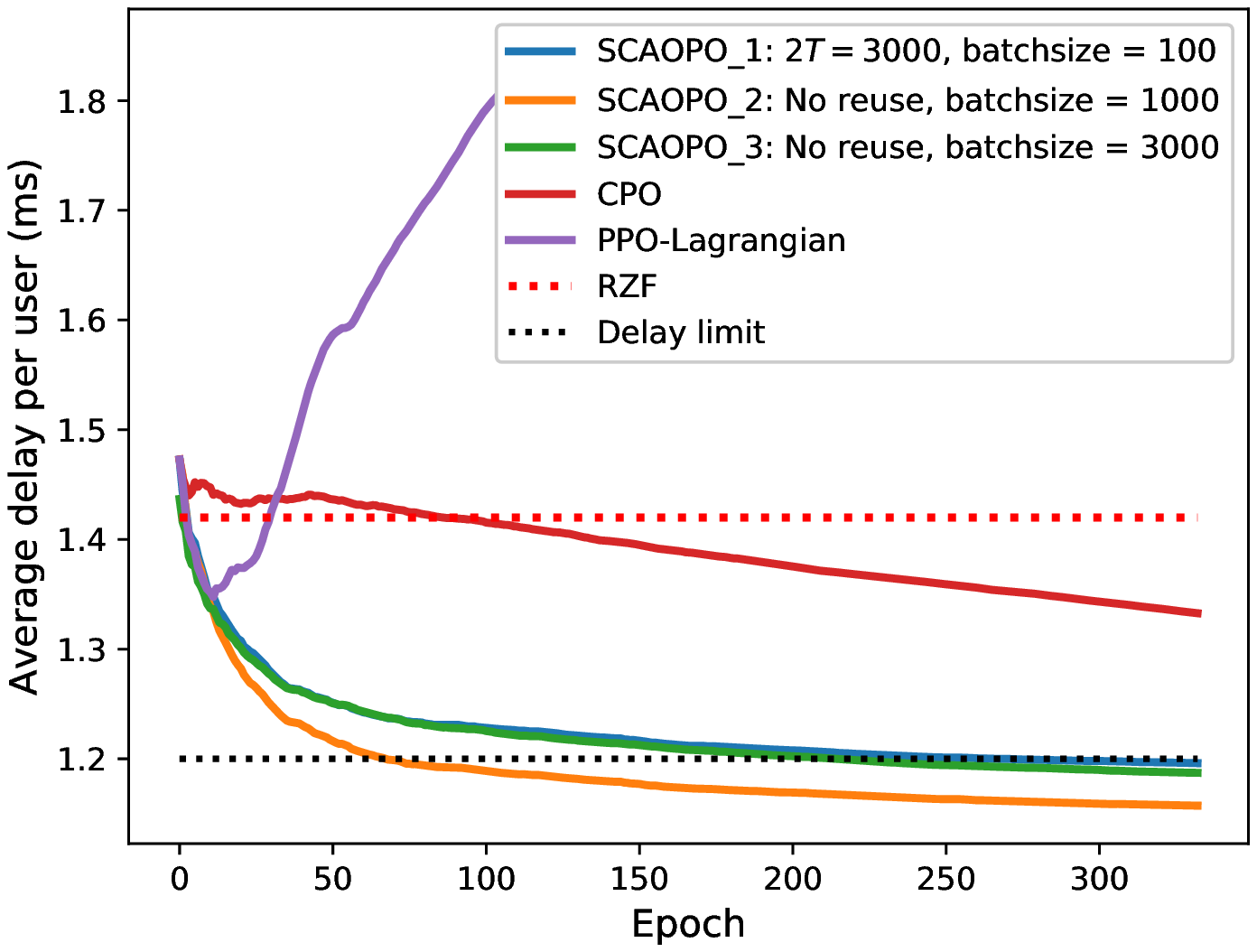}%
		\label{fig: delay0} }
	\caption{ The number of antennas $ N_t = 8 $ and the number of users $ K = 4 $. \textbf{(a)} The learning curve of power consumption. \textbf{(b)} The learning curve of average delay per user.}
	\label{fig: Nt=8}
\end{figure}
\subsection{Delay-Constrained Power Control for Downlink MU-MIMO}\label{sec: sim_example_1}
We adopt a geometry-based channel model as in \cite{2020RCSHP} with a half-wavelength spaced uniform linear array (ULA) for simulations. The channel vector of user $ k $ can be expressed as $ \vect{h}_k=\sum_{i=1}^{N_p}\bar{\alpha}_{k,i}\vect{a}\left(\varphi_{k,i}\right) $,  where $ \vect{a}\roundBrack{\varphi}=\squareBrack{1, e^{j\pi\sin\roundBrack{\varphi}},\dots, e^{j\roundBrack{N_t-1}\pi\sin\roundBrack{\varphi}}}^{\T} $ is the array response vector, $ N_p $ is the number of scattering paths, $ \varphi_{k,i} $'s are angles of departure (AoD) which are Laplacian distributed with an angular spread $ \sigma_{AS}=5 $ and $ \bar{\alpha}_{k,i}\sim\CN{0,\bar{\sigma}^2_{k,i}} $, $ \bar{\sigma}^2_{k,i} $'s are randomly generated from an exponential distribution and normalized such that $ \sum_{i=1}^{N_p}\bar{\sigma}^2_{k,i}=g_k $, and $ g_k $ represents the path gain of user $ k  $. The path gains $ g_k $'s are uniformly generated between -10 dB and 10 dB and the number of scattering paths for each user is  $ N_p=4 $. As similar to \cite{2017Mao_TaskQueue}, we set the bandwith $ B = 10 $ MHz, the duration of one time slot $ \tau = 1 $ ms, the noise power density -100 dBm/Hz and the arrival data rate $ A_k, \forall k $ being uniformly distributed over $ \squareBrack{0, 20} $ Mbit/s. The constants in the surrogate problems are $ \varsigma_k = 1, \forall k $, and the step sizes are $ \alpha_t = \frac{1}{t^{0.6}} $ and $ \beta_t = \frac{1}{t^{0.8}} $.

\figurename~\ref{fig: Nt=8} illustrates the average power consumption and the average delay per user when the number of antennas $ N_t = 8 $ and the number of users $ K = 4 $. In addition to the CRL baselines, we also compare with the RZF precoding with equal power allocation and the asymptotically optimal regularization factor $ \alpha_{Z} = \frac{\sigma^2}{p} $ \cite{2005RZF}, where $ \sigma^2 $ is the noise power and $ p $ is the power equally allocated to each user. It can be seen from \figurename~\ref{fig: Nt=8} that the proposed SCAOPO can significantly reduce the power consumption while meeting the delay constraint, compared with the PPO-Lagrangian and CPO. This reveals the benefit of KKT point convergence guarantee of SCAOPO. Compared with the SCAOPO's without reusing previous experiences (SCAOPO\_2 and SCAOPO\_3), the standard SCAOPO (SCAOPO\_1) can achieve almost the same convergence speed as the case (SCAOPO\_3) where we need to feed much more experiences at each iteration, which indicates that the proposed SCAOPO can attain appealing performance with low complexity.

 \begin{figure}[!t]
	\centering
	\subfloat[]{\includegraphics[width=3in]{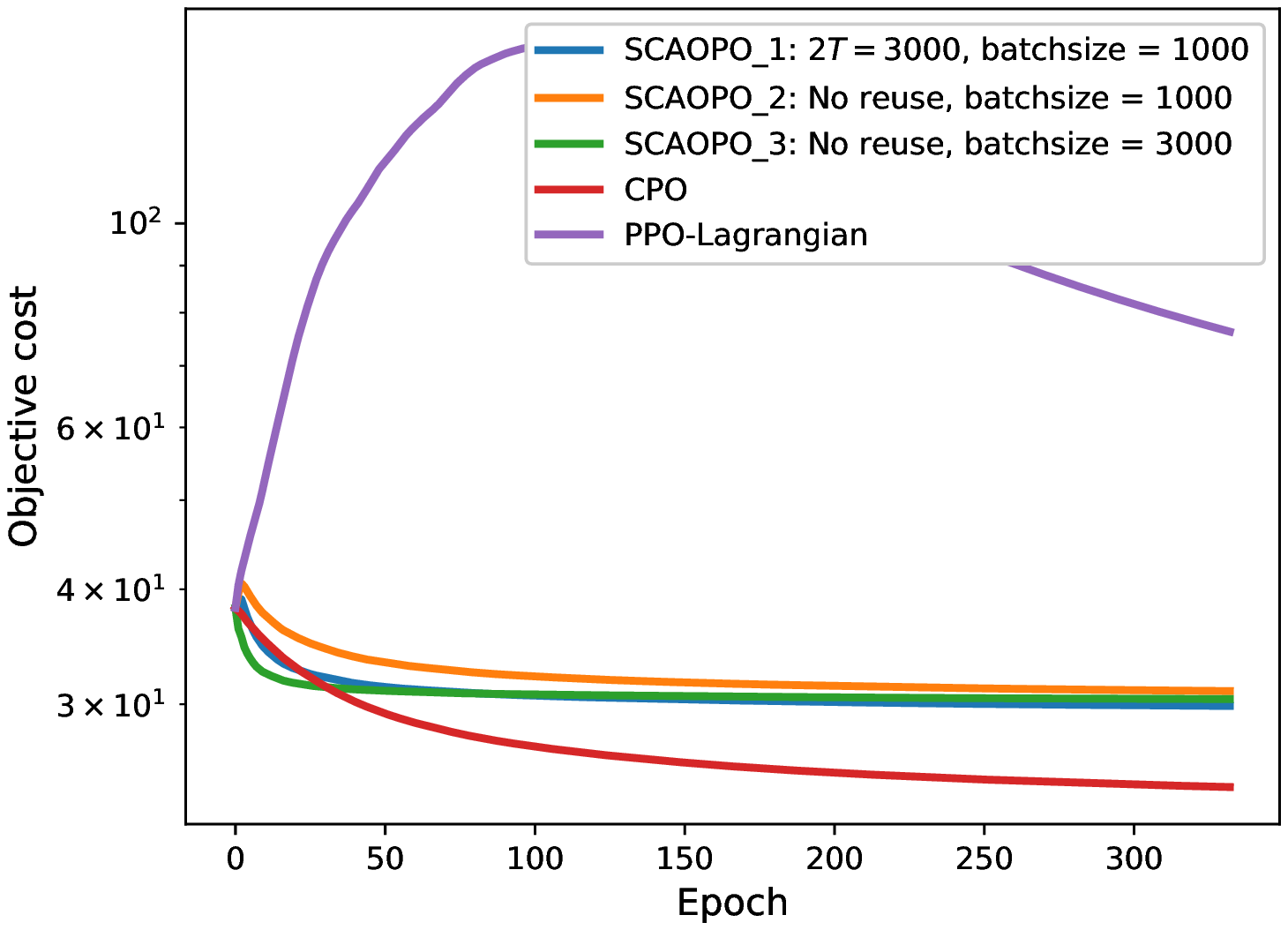}%
		\label{fig: obj_cost_lqr} } 	\hfil
	\subfloat[]{\includegraphics[width=3in]{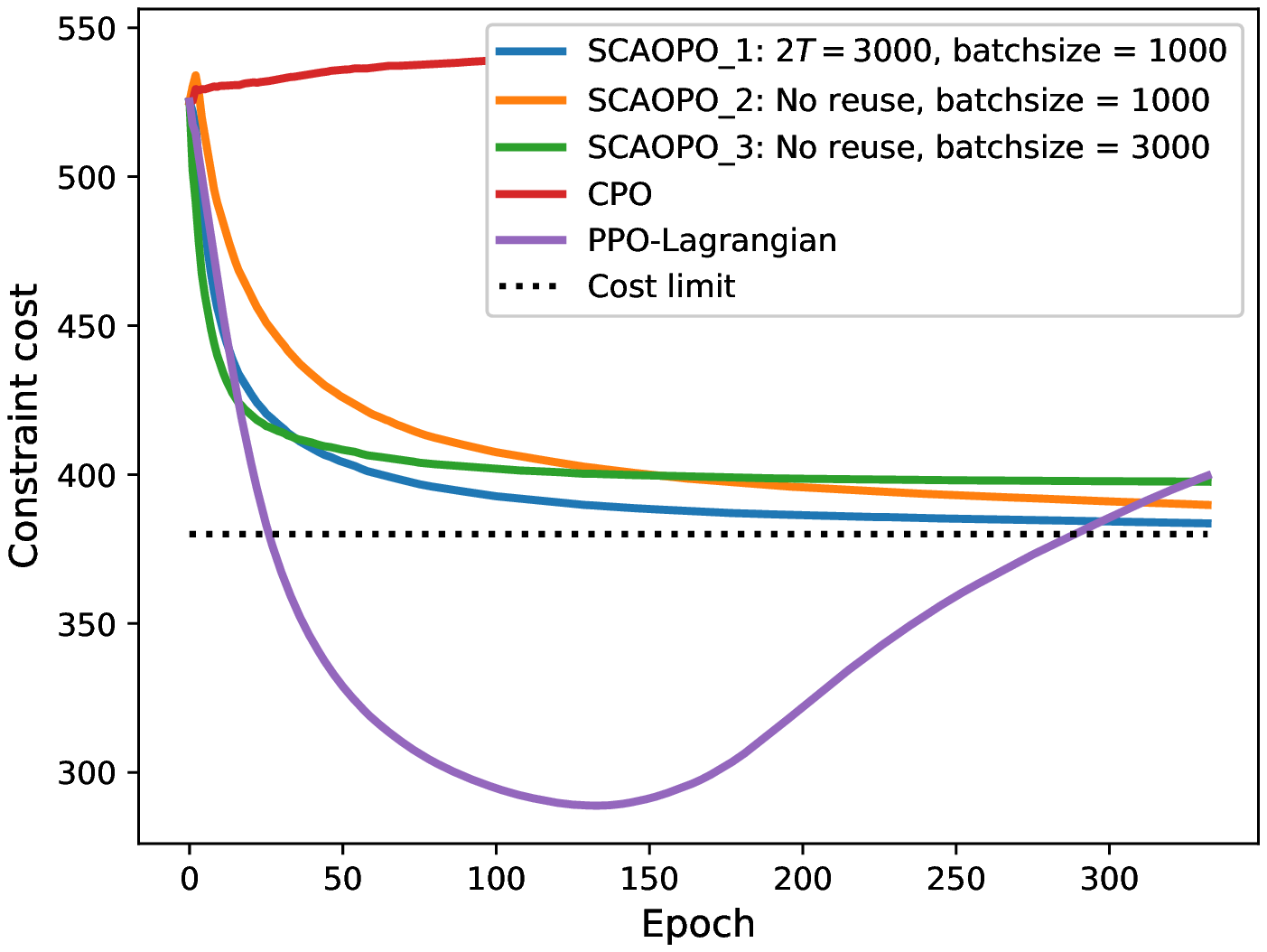}%
		\label{fig: cost_lqr} }
	\caption{\textbf{(a)} The learning curve of objective cost. \textbf{(b)} The learning curve of constraint cost.}
	\label{fig: lqr}
\end{figure}
\subsection{Constrained Linear-Quadratic Regulator} 
As similar to \cite{2019_NIPS_ConvergentPolicyOptimization}, we impose one constraint on the LQR. The state and action dimensions are $ n_s = 15 $ and $ n_a = 4 $, respectively. The transition matrices $ A $ and $ B $ are Gaussian matrices and the cost matrices $ Q_i, \forall i $ and $ R_i, \forall i $ are semi-positive definite matrices. The constants in the surrogate problems are $ \varsigma_i = 10, \forall i $, and the step sizes are $ \alpha_t = \frac{1}{t^{0.6}} $ and $ \beta_t = \frac{1}{t^{0.9}} $.

\figurename~\ref{fig: lqr} demonstrates the learning curves of objective cost and constraint cost. We can observe the similar behavior as in Section \ref{sec: sim_example_1} that the proposed SCAOPO can achieve a significant performance gain over baselines, as well as attaining a good trade-off between performance and complexity.
  
\section{Conclusion}\label{sec: conclusion}
We propose a novel SCAOPO algorithm to solve the general CMDP in the average cost case. At each iteration, SCAOPO updates policy parameter by solving a convex approximation of the original problem if the approximation problem is feasible. Otherwise, the policy parameter is updated by solving the problem of minimizing the constraint violation of the convex approximation. The convex surrogated problems are constructed by off-policy estimation strategy and can be efficiently solved in parallel, which indicates that the SCAOPO enables to dramatically reduce the implementation cost interacting with the environment as well as the computation complexity. Under some technical conditions, we establish the convergence of SCAOPO to a KKT point of the original problem almost surely. Numerical results have demonstrated that the proposed SCAOPO achieve state-of-the-art performance over advanced baselines.

\appendices
\section{Proof of Lemma \ref{lemma: asymptotic consistency}}\label{Proof:Lemma of consistency}
Our proof relies on \cite[Lemma 1]{1980FeasibleDirection}, which is restated below for completeness.
\begin{lemma}\label{lem: stochastic process convergence}
	Let $\left(\Omega,\mathcal{F},\mathbb{P}\right)$ be a probability space and let $\left\{ \mathcal{F}_{t}\right\} $ be an increasing sequence of $\sigma$-field contained in $\mathcal{F}$. Let $\left\{ \eta^{t}\right\} ,\left\{ z^{t}\right\} $ be sequences of $\mathcal{F}_{t}$-measurable random vectors satisfying the relations 
	\begin{align*}
	z^{t+1} & =\Pi_{\mathcal{Z}}\left(z^{t}+\alpha_t\left(\zeta^{t}-z^{t}\right)\right), z^{0}\in\mathcal{Z},\\
	\mathbb{E}\left[\zeta^{t}|\mathcal{F}_{t}\right] & =\eta^{t} + b^{t},
	\end{align*}
	where $\alpha_t\geq0$ and the set $\mathcal{Z}$ is convex and closed, $\Pi_{\mathcal{Z}}\left(\cdot\right)$ denotes projection on $\mathcal{Z}$. Next, let
	
	(a) all accumulation points of the sequence $\left\{ \eta^{t}\right\} $ belong to $\mathcal{Z}$ w.p.1.,
	
	(b) there exists a constant $C$ such that $\mathbb{E}\left[\left\Vert \zeta^{t}\right\Vert^{2}|\mathcal{F}_{t}\right]\leq C$ for all $t\geq0$,
	
	(c) $\sum_{t=0}^{\infty}\mathbb{E}\left[\left(\alpha_{t}\right)^{2}+\alpha_t\left\Vert b^{t}\right\Vert \right]<\infty$,
	
	(d) $\sum_{t=0}^{\infty}\alpha_t=\infty$, and (e)
		$\left\Vert \eta^{t+1}-\eta^{t}\right\Vert /\alpha_{t}\rightarrow0$ w.p.1.
	
	Then $z^{t}-\eta^{t}\rightarrow0$ w.p.1.
\end{lemma}
Based on Lemma \ref{lem: stochastic process convergence}, we prove the asymptotic consistency of function value \eqref{eqn: function value consistency} and gradient \eqref{eqn: gradient consistency}, respectively.

\subsection{Proof of \eqref{eqn: function value consistency}}\label{sec: value consistency proof}
Since the cost function $ C_i\roundBrack{s,a} $ is bounded and the step size $ \curlyBrack{\alpha_t} $ satisfies the Assumption \ref{Assmp: stepsizes}, it can be easily verified that the technical conditions (a), (b) and (d) in Lemma \ref{lem: stochastic process convergence} are satisfied. For the condition (c), we have the stochastic bias at time $ t $ as 
\begin{align}
\norm{b^t_i} &= \abs{\Expect\squareBrack{\tilde{J}^t_i} - J_i\roundBrack{\theta_t}} \notag\\
&\leq\abs{\Expect\squareBrack{\tilde{J}^t_i} - \Expect\squareBrack{\tilde{\bar{J}}^t_i}} + \abs{\Expect\squareBrack{\tilde{\bar{J}}^t_i} - J_i\roundBrack{\theta_t}} , \forall i,	
\end{align}
where the inequality follows from the triangle inequality, and $ \tilde{\bar{J}}^t_i $ is defined by the same form as \eqref{eqn: function value estimate} but using the auxiliary experiences $ \tilde{\varepsilon}_t = \Big\{\tilde{s}_{t-2T+1}, \tilde{a}_{t-2T+1}, \curlyBrack{C^{'}_i\roundBrack{\tilde{s}_{t-2T+1}, \tilde{a}_{t-2T+1}}}_{i = 0,\dots,m}, \tilde{s}_{t-2T+2}, \\
\dots, \tilde{s}_t, \tilde{a}_t, \curlyBrack{C^{'}_i\roundBrack{\tilde{s}_t,\tilde{a}_t}}_{i = 0,\dots,m}\Big\} $, which is obtained from an auxiliary trajectory $ \tilde{p_s} $ which is used to assist proof:
\begin{align}
\tilde{p_s} =& \Big\{\tilde{s}_0, \tilde{a}_0, \tilde{s}_1, \dots, \tilde{s}_{t-2T-\tau+1}, \tilde{a}_{t-2T-\tau+1}, \tilde{s}_{t-2T-\tau+2}, \dots, \notag\\ 
&\quad  \tilde{s}_{t-2T+1}, \tilde{a}_{t-2T+1}, \tilde{s}_{t-2T+2}, \dots, \tilde{s}_t, \tilde{a}_t\Big\}.\notag
\end{align} 
Before time $ t-2T-\tau $, the states and actions are generated according to time-varying polices as $ p_s $ of Problem \eqref{original_problem}, but after time $ t-2T-\tau $, the policy is kept fixed as $ \pi_{\theta^{t-2T-\tau}} $ to generate all the subsequent actions, i.e., $ a_{l}\sim\pi_{\theta^{l}}$ when $ l=0,1,\dots, t-2T-\tau$ and $ a_{l}\sim\pi_{\theta^{t-2T-\tau}}$ when $ l=t-2T-\tau+1, \dots, t$. 

According to the definitions, we have
\begin{align}\label{eqn: J_tilde}
\Expect\squareBrack{\tilde{J}^t_i} = \frac{1}{2T}\sum_{l=1}^{2T}& \int_{s\in\mathcal{S}}\mathbf{P}\roundBrack{S_{t-2T+l}=\mathrm{d}s}, \notag \\ &\int_{a\in\mathcal{A}}\pi_{\theta^{t-2T+l}}\roundBrack{\mathrm{d}a|s}C^{'}_i\roundBrack{s,a},  
\end{align}
\begin{align}\label{eqn: J_bar_tilde}
\Expect\squareBrack{\tilde{\bar{J}}^t_i} = \frac{1}{2T}\sum_{l=1}^{2T}& \int_{s\in\mathcal{S}}\mathbf{P}\roundBrack{\tilde{S}_{t-2T+l}=\mathrm{d}s}, \notag \\ &\int_{a\in\mathcal{A}}\pi_{\theta^{t-2T-\tau}}\roundBrack{\mathrm{d}a|s}C^{'}_i\roundBrack{s,a}, 
\end{align}
\begin{equation}\label{eqn: J}
	J_i\roundBrack{\theta_t} = \int_{s\in\mathcal{S}}\mathbf{P}_{\pi_{\theta^t}}\roundBrack{\mathrm{d}s} \int_{a\in\mathcal{A}}\pi_{\theta^t}\roundBrack{\mathrm{d}a|s}C^{'}_i\roundBrack{s,a} ,
\end{equation}
where \eqref{eqn: J} follows from the ergodicity assumption in Assumption \ref{Problem Assumptions}. Then
\begin{align}\label{eqn:bias1_function}
&\abs{\Expect\squareBrack{\tilde{J}^t_i} - \Expect\squareBrack{\tilde{\bar{J}}^t_i}} \notag\\
\leq&\frac{1}{2T}\sum_{l=1}^{2T}\bigg|\int_{s\in\mathcal{S}}\mathbf{P}\roundBrack{S_{t-2T+l}=\mathrm{d}s} \int_{a\in\mathcal{A}}\pi_{\theta^{t-2T+l}}\roundBrack{\mathrm{d}a|s}C^{'}_i\roundBrack{s,a}\notag\\
&-\int_{s\in\mathcal{S}}\mathbf{P}\roundBrack{\tilde{S}_{t-2T+l}=\mathrm{d}s} \int_{a\in\mathcal{A}}\pi_{\theta^{t-2T+l}}\roundBrack{\mathrm{d}a|s}C^{'}_i\roundBrack{s,a}\notag\\
&+ \int_{s\in\mathcal{S}}\mathbf{P}\roundBrack{\tilde{S}_{t-2T+l}=\mathrm{d}s} \int_{a\in\mathcal{A}}\pi_{\theta^{t-2T+l}}\roundBrack{\mathrm{d}a|s}C^{'}_i\roundBrack{s,a}\notag\\
&-\int_{s\in\mathcal{S}}\mathbf{P}\roundBrack{\tilde{S}_{t-2T+l}=\mathrm{d}s} \int_{a\in\mathcal{A}}\pi_{\theta^{t-2T-\tau}}\roundBrack{\mathrm{d}a|s}C^{'}_i\roundBrack{s,a}\bigg|\notag\\
\overset{a}{\leq}&\frac{1}{2T}\sum_{l=1}^{2T}\int_{s\in\mathcal{S}}\abs{\mathbf{P}\roundBrack{S_{t-2T+l}=\mathrm{d}s} - \mathbf{P}\roundBrack{\tilde{S}_{t-2T+l}=\mathrm{d}s}}\notag\\
&\int_{a\in\mathcal{A}}\pi_{\theta^{t-2T+l}}\roundBrack{\mathrm{d}a|s}\abs{C^{'}_i\roundBrack{s,a}}\notag\\
&+\frac{1}{2T}\sum_{l=1}^{2T}\int_{s\in\mathcal{S}} \mathbf{P}\roundBrack{\tilde{S}_{t-2T+l}=\mathrm{d}s}\notag\\
&\int_{a\in\mathcal{A}}\abs{\pi_{\theta^{t-2T+l}}\roundBrack{\mathrm{d}a|s} - \pi_{\theta^{t-2T-\tau}}\roundBrack{\mathrm{d}a|s}}\abs{C^{'}_i\roundBrack{s,a}}\notag\\
=&\frac{1}{2T}\sum_{l=1}^{2T}O\bigg(\norm{\mathbf{P}\roundBrack{S_{t-2T+l}\in\cdot} - \mathbf{P}\roundBrack{\tilde{S}_{t-2T+l}\in\cdot}}_{TV} \notag\\
&+ \norm{\theta^{t-2T+l} - \theta^{t-2T-\tau}}\bigg), 
\end{align}
where \eqref{eqn:bias1_function}-a follows from triangle inequality and Cauchy-Schwarz inequality, and the last equality comes from the bounded action space, the bounded cost and the Lipschitz continuity of the policy $ \pi_{\theta} $ over the parameter $ \theta $. Moreover, according to the update rule of $ \theta $ in \eqref{eqn: policy update} and the compactness of $ \Theta $, we have
\begin{equation}\label{eqn: step_size_order}
	\norm{\theta^{t-2T+l} - \theta^{t-2T-\tau}}= O\roundBrack{\sum_{t^{'}=t-2T-\tau+1}^{t-2T+l}\beta_{t^{'}}}.
\end{equation}

Define the transition kernel $ K_{\theta}\roundBrack{s,\mathrm{d}s^{'}} $ as
\begin{equation}\label{eqn: transition kernel}
K_{\theta}\roundBrack{s,\mathrm{d}s^{'}} = \int_{a\in\mathcal{A}}P\roundBrack{\mathrm{d}s^{'}|s,a}\pi_{\theta}\roundBrack{\mathrm{d}a|s},
\end{equation}
which refers to the probability of the state $ s $ transitioning to the state $ s^{'} $ under the policy $ \pi_{\theta} $. Then
\begin{align}
&\norm{\mathbf{P}\roundBrack{S_{t-2T+l}\in\cdot} - \mathbf{P}\roundBrack{\tilde{S}_{t-2T+l}\in\cdot}}_{TV}\notag\\
=& \int_{s^{'}\in\mathcal{S}}\abs{\mathbf{P}\roundBrack{S_{t-2T+l}=\mathrm{d}s^{'}} - \mathbf{P}\roundBrack{\tilde{S}_{t-2T+l}=\mathrm{d}s^{'}}} \notag \\
=& \int_{s^{'}\in\mathcal{S}}\bigg| \int_{s\in\mathcal{S}}\mathbf{P}\roundBrack{S_{t-2T+l-1}=\mathrm{d}s}K_{\theta^{t-2T+l-1}}\roundBrack{s, \mathrm{d}s^{'}}\notag \\
&-\int_{s\in\mathcal{S}}\mathbf{P}\roundBrack{\tilde{S}_{t-2T+l-1}=\mathrm{d}s}K_{\theta^{t-2T-\tau}}\roundBrack{s, \mathrm{d}s^{'}} \bigg| \notag
\end{align}
\begin{align}\label{eqn:recursive_TV_explansion}
\overset{a}{\leq}&\int_{s^{'}\in\mathcal{S}}\int_{s\in\mathcal{S}}\abs{\mathbf{P}\roundBrack{S_{t-2T+l-1}=\mathrm{d}s} - \mathbf{P}\roundBrack{\tilde{S}_{t-2T+l-1}=\mathrm{d}s}}\notag\\
&\qquad\qquad K_{\theta^{t-2T+l-1}}\roundBrack{s, \mathrm{d}s^{'}}\notag\\
&+\int_{s^{'}\in\mathcal{S}}\int_{s\in\mathcal{S}}\mathbf{P}\roundBrack{\tilde{S}_{t-2T+l-1}=\mathrm{d}s}\notag\\
&\quad \abs{K_{\theta^{t-2T+l-1}}\roundBrack{s, \mathrm{d}s^{'}} - K_{\theta^{t-2T-\tau}}\roundBrack{s, \mathrm{d}s^{'}}},
\end{align}
where \eqref{eqn:recursive_TV_explansion}-$ a $ follows from the similar tricks as \eqref{eqn:bias1_function}-a, i.e., minus and add intermediate terms while using triangle inequality and Cauchy-Schwarz inequality, and
\begin{align}
	&\mbox{The first term} = \norm{\mathbf{P}\roundBrack{S_{t-2T+l-1}\in\cdot} - \mathbf{P}\roundBrack{\tilde{S}_{t-2T+l-1}=\in\cdot}}_{TV} \\
	&\mbox{The second term} = O\roundBrack{\sum_{t^{'}=t-2T-\tau+1}^{t-2T+l-1}\beta_{t^{'}}} \label{eqn:M2}.
\end{align}
The second term follows from expanding the transition kernel by the definition in \eqref{eqn: transition kernel} and applying the Lipschitz continuity of the policy $ \pi_{\theta} $ over the parameter $ \theta $, together with the boundness of action space and cost function and the result of \eqref{eqn: step_size_order}. Then we have
\begin{align}\label{eqn: one step expanding}
&\norm{\mathbf{P}\roundBrack{S_{t-2T+l}\in\cdot} - \mathbf{P}\roundBrack{\tilde{S}_{t-2T+l}\in\cdot}}_{TV}\notag\\
\leq&\norm{\mathbf{P}\roundBrack{S_{t-2T+l-1}\in\cdot} - \mathbf{P}\roundBrack{\tilde{S}_{t-2T+l-1}\in\cdot}}_{TV} \notag\\ 
&+ O\roundBrack{\sum_{t^{'}=t-2T-\tau+1}^{t-2T+l-1}\beta_{t^{'}}}.
\end{align}

According to the definition of the auxiliary trajectory $ \tilde{p_s} $, we have $ \mathbf{P}\roundBrack{S_{t-2T-\tau+1}\in\cdot} = \mathbf{P}\roundBrack{\tilde{S}_{t-2T-\tau+1}\in\cdot} $. By recursively using \eqref{eqn: one step expanding}, we obtain
\begin{align}\label{eqn: T steps expanding}
&\norm{\mathbf{P}\roundBrack{S_{t-2T+l}\in\cdot} - \mathbf{P}\roundBrack{\tilde{S}_{t-2T+l}\in\cdot}}_{TV}\notag\\
=&O\roundBrack{\sum_{j=t-2T-\tau+1}^{t-2T+l-1}\sum_{t^{'}=t-2T-\tau+1}^{j}\beta_{t^{'}}}.
\end{align}
Meanwhile,
\begin{align}\label{eqn: finnal_expansion}
&\sum_{j=t-2T-\tau+1}^{t-2T+l-1}\sum_{t^{'}=t-2T-\tau+1}^{j}\beta_{t^{'}}\leq\sum_{j=t-2T-\tau+1}^{t-2T+l-1}\sum_{t^{'}=t-2T-\tau+1}^{t-2T+l-1}\beta_{t^{'}} \notag\\
=&\roundBrack{l-1+\tau}\sum_{t^{'}=t-2T-\tau+1}^{t-2T+l-1}\beta_{t^{'}}\leq\roundBrack{T-1+\tau}\sum_{t^{'}=t-2T-\tau+1}^{t}\beta_{t^{'}}.
\end{align}
Combing \eqref{eqn:bias1_function} with \eqref{eqn: step_size_order}, \eqref{eqn: T steps expanding} and \eqref{eqn: finnal_expansion}, we have
\begin{equation}\label{eqn:bias1_function_final}
	\abs{\Expect\squareBrack{\tilde{J}^t_i} - \Expect\squareBrack{\tilde{\bar{J}}^t_i}} = O\roundBrack{\roundBrack{T+\tau}\sum_{t^{'}=t-2T-\tau+1}^{t}\beta_{t^{'}}}.
\end{equation}

Use the definitions in \eqref{eqn: J_bar_tilde} and \eqref{eqn: J} and apply the similar tricks as in \eqref{eqn:bias1_function}, we have
\begin{align}\label{eqn:bias2_function}
&\abs{\Expect\squareBrack{\tilde{\bar{J}}^t_i} - J_i\roundBrack{\theta^t}} \notag\\
=&\frac{1}{2T}\sum_{l=1}^{2T}O\bigg(\norm{\mathbf{P}\roundBrack{\tilde{S}_{t-2T+l}\in\cdot} - \mathbf{P}_{\pi_{\theta_t}}}_{TV}+ \norm{\theta^{t} - \theta^{t-2T-\tau}}\bigg),
\end{align}
where 
\begin{align}\label{eqn:distribution_bias}
&\norm{\mathbf{P}\roundBrack{\tilde{S}_{t-2T+l}\in\cdot} - \mathbf{P}_{\pi_{\theta^t}}}_{TV} \notag\\
\leq&\norm{\mathbf{P}\roundBrack{\tilde{S}_{t-2T+l}\in\cdot} - \mathbf{P}_{\pi_{\theta^{t-2T-\tau}}}}_{TV} + \norm{\mathbf{P}_{\pi_{\theta^{t-2T-\tau}}} - \mathbf{P}_{\pi_{\theta^t}}}_{TV} \notag \\
\overset{a}{\leq}&m_c\rho^{\tau+l} + \roundBrack{\lceil\log_{\rho}^{m_c^{-1}}\rceil + \frac{1}{1-\rho}}\norm{K_{\theta^{t}} - K_{\theta^{t-2T-\tau}}}_{q},
\end{align}
where \eqref{eqn:distribution_bias}-$ a $ follows from the ergodicity assumption (the first term) and the Theorem 3.1 in \cite{2005Convergence_Ergodic_MDP} (the second term), and $ \norm{\cdot}_q $ is the operator norm: $ \norm{A}_q\doteq\sup_{\norm{q}_{TV}=1}\norm{qA}_{TV} $. Then
\begin{align}\label{eqn:transtion_kernel_difference}
&\norm{K_{\theta^{t}} - K_{\theta^{t-2T-\tau}}}_q \notag\\
=&\sup_{\norm{q}_{TV}=1}\norm{ \int_{s\in\mathcal{S}}q\roundBrack{\mathrm{d}s}\roundBrack{K_{\theta^{t}} - K_{\theta^{t-2T-\tau}}}\roundBrack{s, \cdot} }_{TV} \notag\\
=&\sup_{\norm{q}_{TV}=1}\int_{s^{'}\in\mathcal{S}}\abs{\int_{s\in\mathcal{S}}q\roundBrack{\mathrm{d}s}\roundBrack{K_{\theta^{t}} - K_{\theta^{t-2T-\tau}}}\roundBrack{s, \mathrm{d}s^{'}}}\notag\\
\leq&\sup_{\norm{q}_{TV}=1}\int_{s^{'}\in\mathcal{S}}\int_{s\in\mathcal{S}}\abs{q\roundBrack{\mathrm{d}s}}\abs{K_{\theta^{t}}\roundBrack{s, \mathrm{d}s^{'}} - K_{\theta^{t-2T-\tau}}\roundBrack{s, \mathrm{d}s^{'}}} \notag\\
\overset{a}{=}&O\roundBrack{\sum_{t^{'}=t-2T-\tau+1}^{t}\beta_{t^{'}}},
\end{align}
where \eqref{eqn:transtion_kernel_difference}-$ a $ follows the same tricks as in \eqref{eqn:M2}. Combining \eqref{eqn:bias2_function} with \eqref{eqn:distribution_bias}, \eqref{eqn:transtion_kernel_difference} as well as $ \rho^{\tau+l}\leq\rho^{\tau} $, we have
\begin{equation}\label{eqn:bias2_function_final}
\abs{\Expect\squareBrack{\tilde{\bar{J}}^t_i} - J_i\roundBrack{\theta^t}} = O\roundBrack{\rho^{\tau} + \sum_{t^{'}=t-2T-\tau+1}^{t}\beta_{t^{'}}}
\end{equation}

In summary, we obtain the stochastic bias of estimating function value at time $ t $ by combing \eqref{eqn:bias1_function_final} with \eqref{eqn:bias2_function_final}:
\begin{equation}
\norm{b_i^t} = O\roundBrack{ \rho^{\tau} + \roundBrack{T+\tau}\sum_{t^{'}=t-2T-\tau+1}^{t}\beta_{t^{'}} }
\end{equation}
Let $ \rho^{\tau} = O\roundBrack{\frac{1}{t}} $ which means $ \tau = O\roundBrack{\log t} $. Then according to the assumption of step size in Assumption \ref{Assmp: stepsizes}-1), we have $ \sum_t\alpha_t \norm{b_i^t} < \infty $, thus the technical condition (c) in Lemma \ref{lem: stochastic process convergence} is satisfied.

For the technical condition (e), according to the definition in \eqref{eqn: J}, we have 
\begin{align}\label{eqn:eta_function}
\abs{J\roundBrack{\theta_t} - J\roundBrack{\theta_{t-1}}} \overset{a}{=}& O\roundBrack{\norm{\mathbf{P}_{\pi_{\theta^t}} - \mathbf{P}_{\pi_{\theta^{t-1}}}}_{TV} +  \norm{\theta^{t} - \theta^{t-1}}} \notag\\
\overset{b}{=}&O\roundBrack{\norm{\theta^{t} - \theta^{t-1}}} = O\roundBrack{\beta_t},
\end{align}
where \eqref{eqn:eta_function}-$ a $ follows from the similar tricks as in \eqref{eqn:bias1_function} and \eqref{eqn:eta_function}-$ b $ follows from the Theorem 3.1 in \cite{2005Convergence_Ergodic_MDP} and the result of \eqref{eqn:transtion_kernel_difference}. It can be seen from the Assumption \ref{Assmp: stepsizes}-3) that technical condition (e) is satisfied. This completes the proof of \eqref{eqn: function value consistency}.

\subsection{Proof of \eqref{eqn: gradient consistency}}
We first construct an auxiliary gradient estimate:
\begin{equation}\label{eqn: auxiliary gradient moving_avg}
\hat{\bar{g}}^t_i = \roundBrack{1-\alpha_t}\hat{\bar{g}}^{t-1}_i + \alpha_t\tilde{\bar{g}}^t_i,
\end{equation}
where 
\begin{align}\label{eqn: gradient estimate auxiliary}
\tilde{\bar{g}}_i^t = \frac{1}{T}\sum_{l=1}^{T}&\hat{\bar{Q}}_i^{t-2T+l}\roundBrack{s_{t-2T+l}, a_{t-2T+l}}\notag\\ &\nabla\log\pi_{\theta^t}\roundBrack{a_{t-2T+l}|s_{t-2T+l}},
\end{align}
where 
\begin{align}
\hat{\bar{Q}}_i^{t-2T+l}&\roundBrack{s_{t-2T+l}, a_{t-2T+l}} = \notag\\
&\sum_{l^{'} = 0}^{T-1}\roundBrack{ C^{'}_i\roundBrack{s_{t-2T+l+l^{'}}, a_{t-2T+l+l^{'}}} - J_i\roundBrack{\theta^t} }. \notag
\end{align}
Note that the only difference between \eqref{eqn: gradient estimate} and \eqref{eqn: gradient estimate auxiliary} comes from replacing the term $ \hat{J}_i^t $ with $ J_i\roundBrack{\theta^t} $ in the estimate of Q-value. We use two steps to prove the asymptotic consistency of gradient \eqref{eqn: gradient consistency}:
\begin{align}
\mbox{Step 1:}\  &\lim_{t\to\infty} \norm{\hat{\bar{g}}^t_i - \nabla J_i\roundBrack{\theta^t}}_2 = 0 \label{eqn: step 1}\\
\mbox{Step 2:}\  &\lim_{t\to\infty} \norm{\hat{g}^t_i - \hat{\bar{g}}^t_i}_2 = 0 \label{eqn: step 2}
\end{align}

\subsubsection{Step 1: Proof of \eqref{eqn: step 1}}\label{sec: step 1}
It can be verified that the technical conditions (a), (b), (d) and (e) in Lemma \ref{lem: stochastic process convergence} are satisfied for the auxiliary gradient estimate \eqref{eqn: gradient estimate auxiliary} following the similar analysis as in Appendix \ref{sec: value consistency proof}. Moreover, we have 
\begin{align}
	\Expect\squareBrack{\tilde{\bar{g}}_i^t} &= \frac{1}{T}\sum_{l=1}^{T}\int_{s\in\mathcal{S}}\mathbf{P}\roundBrack{S_{t-2T+l}=\mathrm{d}s} \int_{a\in\mathcal{A}}\pi_{\theta^{t-2T+l}}\roundBrack{\mathrm{d}a|s}\notag\\
	&\qquad\qquad\Expect\squareBrack{\hat{\bar{Q}}^{t-2T+l}_i\roundBrack{s, a}}\nabla\log\pi_{\theta^t}\roundBrack{a|s}
\end{align}
where we use a simplified notation $ \Expect\squareBrack{\hat{\bar{Q}}^{t-2T+l}_i\roundBrack{s, a}} = \Expect\squareBrack{\hat{\bar{Q}}^{t-2T+l}_i\roundBrack{s, a} \Big| S_{t-2T+l}=s, A_{t-2T+l}=a} $ for conciseness. Together with the definition of real gradient in \eqref{eqn: policy gradient}, for technical condition (c), we have 
\begin{align}\label{eqn:bias_gradient}
	&\norm{\Expect\squareBrack{\tilde{\bar{g}}_i^t} - \nabla J_i\roundBrack{\theta^t}}_2\notag\\
	\overset{a}{=}&\frac{1}{T}\sum_{l=1}^{T}O\bigg(\norm{\mathbf{P}\roundBrack{S_{t-2T+l}\in\cdot} - \mathbf{P}_{\pi_{\theta^{t}}}}_{TV}+ \norm{\theta^{t} - \theta^{t-2T+l}}_2 \notag\\
	 &\qquad+ \abs{\Expect\squareBrack{\hat{\bar{Q}}^{t-2T+l}_i\roundBrack{s, a}} - Q_{\pi_{\theta^{t}}}\roundBrack{s,a}}\bigg),
\end{align}
where \eqref{eqn:bias_gradient}-$ a $ follows from similar tricks as in \eqref{eqn:bias1_function}, the Lipschitz continuity of policy $ \pi_{\theta} $ over the parameter $ \theta $ and the regularity conditions, i.e., $ \forall (s, a) $, $ C_i\roundBrack{s,a} $ and $ \nabla\log\pi_{\theta}\roundBrack{a|s} $ are bounded.

Following from the similar analysis as in \eqref{eqn:recursive_TV_explansion} and \eqref{eqn:distribution_bias},
we have 
\begin{align}\label{eqn:distribution_bias1}
		&O\roundBrack{\norm{\mathbf{P}\roundBrack{S_{t-2T+l}\in\cdot} - \mathbf{P}_{\pi_{\theta^{t}}}}_{TV}}\notag\\
		 \leq& O\roundBrack{ \rho^{\tau} + \roundBrack{T+\tau}\sum_{t^{'}=t-2T-\tau+1}^{t}\beta_{t^{'}} }
\end{align}
Moreover,
\begin{align}
	&\Expect\squareBrack{\hat{\bar{Q}}^{t-2T+l}_i\roundBrack{s, a}}\notag\\
	=& C^{'}_i\roundBrack{s,a} - J_i\roundBrack{\theta^{t}} + \int_{\mathcal{S}}P\roundBrack{\mathrm{d}s^{'}|s,a}\notag\\
	&\qquad\int_{\mathcal{A}}\pi_{\theta^{t-2T+l+1}}\roundBrack{\mathrm{d}a^{'}|s^{'}}C^{'}_i\roundBrack{s^{'}, a^{'}} - J_i\roundBrack{\theta_{t}} \notag\\
	&+ \sum_{l^{'} = 2}^{T-1}\bigg(\int_{\mathcal{S}}\mathbf{P}\roundBrack{S_{t-2T+l+l^{'}}=\mathrm{d}s^{'}} \int_{\mathcal{A}}\pi_{\theta^{t-2T+l+l^{'}}}\roundBrack{\mathrm{d}a^{'}|s^{'}}\notag\\
	&\qquad\qquad C^{'}_i\roundBrack{s^{'}, a^{'}} - J_i\roundBrack{\theta^{t}}\bigg).
\end{align}
Together with the definition of real Q-value in \eqref{eqn: Q-value}, we have
\begin{align}\label{eqn:Q_value_bias}
	&\abs{\Expect\squareBrack{\hat{\bar{Q}}^{t-2T+l}_i\roundBrack{s, a}} - Q_i^{\pi_{\theta^t}}\roundBrack{s,a}}\notag\\
	\overset{a}{=}&O\bigg(\norm{\theta^{t} - \theta^{t-2T+l+1}}_2 + \sum_{l^{'} = 2}^{T-1}\Big(\big\lVert\mathbf{P}\roundBrack{S_{t-2T+l+l^{'}}\in\cdot} - \notag\\ &\mathbf{P}_t\roundBrack{S_{l^{'}}\in\cdot}\big\rVert_{TV} + \norm{\theta^{t} - \theta^{t-2T+l+l^{'}}}_2\Big) + \abs{M_3} \bigg),
\end{align}
where \eqref{eqn:Q_value_bias}-$ a $ uses similar tricks as in \eqref{eqn:bias1_function} while applying the Lipschitz continuity of policy $ \pi_{\theta} $ over the parameter $ \theta $, $ \mathbf{P}_t\roundBrack{S_{l^{'}}\in\cdot} $ denotes the state distribution at time $ l^{'} $ under the policy $ \pi_{\theta^t} $, and 
\begin{align}\label{eqn: M3}
\abs{M_3} =& \bigg|\sum_{l^{'} = T}^{\infty}\bigg(\int_{\mathcal{S}}\mathbf{P}_t\roundBrack{S_{l^{'}}=\mathrm{d}s^{'}} \int_{\mathcal{A}}\pi_{\theta^{t}}\roundBrack{\mathrm{d}a^{'}|s^{'}}C^{'}_i\roundBrack{s^{'}, a^{'}} \notag\\ 
&- \int_{\mathcal{S}}\mathbf{P}_{\pi_{\theta^{t}}}\roundBrack{\mathrm{d}s^{'}} \int_{\mathcal{A}}\pi_{\theta^{t}}\roundBrack{\mathrm{d}a^{'}|s^{'}}C^{'}_i\roundBrack{s^{'}, a^{'}} \bigg)\bigg|\notag\\
=& \sum_{l^{'} = T}^{\infty} O\roundBrack{\norm{\mathbf{P}_t\roundBrack{S_{l^{'}}\in\cdot} - \mathbf{P}_{\pi_{\theta_{t}}} }_{TV}}\overset{a}{=} O\roundBrack{\rho^T},
\end{align}
where \eqref{eqn: M3}-$ a $ follows from the ergodicity assumption in Assumption \ref{Problem Assumptions}. According to the similar analysis as in \eqref{eqn:recursive_TV_explansion} and the fact $ \mathbf{P}\roundBrack{S_{t-2T+l+1}\in\cdot} = \mathbf{P}_t\roundBrack{S_{1}\in\cdot} = P\roundBrack{\cdot|s,a} $, we have
\begin{align}\label{eqn:distribution_bias2}
&\norm{\mathbf{P}\roundBrack{S_{t-2T+l+l^{'}}\in\cdot} - \mathbf{P}_t\roundBrack{S_{l^{'}}\in\cdot}}_{TV}\notag\\ 
=& O\roundBrack{\sum_{j=t-2T+l+1}^{t-2T+l+l^{'}}\sum_{t^{'}=j}^{t}\beta_{t^{'}}}
\leq O\roundBrack{T\sum_{t^{'}=t-2T-\tau}^{t}\beta_{t^{'}}}	
\end{align}

Finally, we obtain the stochastic bias of auxiliary gradient estimate by combing \eqref{eqn:bias_gradient} with \eqref{eqn:distribution_bias1}, \eqref{eqn:Q_value_bias}, \eqref{eqn: M3} and \eqref{eqn:distribution_bias2}:
\begin{align}
&\norm{b_i^t} = \norm{\Expect\squareBrack{\tilde{\bar{g}}_i^t} - \nabla J_i\roundBrack{\theta^t}}_2\notag\\
=&O\bigg(\rho^{\tau} + \roundBrack{T+\tau}\sum_{t^{'}=t-2T-\tau+1}^{t}\beta_{t^{'}} + \rho^T + T^2\sum_{t^{'}=t-2T-\tau+1}^{t}\beta_t^{'} \bigg).
\end{align}
Let $ \rho^{\tau} = O\roundBrack{\frac{1}{t}} $ and  $ \rho^{T} = O\roundBrack{\frac{1}{t}} $ which means $ \tau = O\roundBrack{\log t} $ and $ T = O\roundBrack{\log t} $. Then according to the assumption of step size in Assumption \ref{Assmp: stepsizes}-1), we have $ \sum_t\alpha_t \norm{b_i^t} < \infty $, thus the technical condition (c) in Lemma \ref{lem: stochastic process convergence} is satisfied. This completes the proof of \eqref{eqn: step 1}.

\subsubsection{Step 2: Proof of \eqref{eqn: step 2}}\label{sec: step 2}
According to the definitions of \eqref{eqn: gradient_moving_avg} and \eqref{eqn: auxiliary gradient moving_avg}, we have
\begin{align}
&\hat{g}^t_i = \sum_{t^{'}=0}^{t}\prod_{j=t^{'}+1}^{t}\roundBrack{1-\alpha_j}\alpha_{t^{'}} \tilde{g}^{t^{'}}_i \notag\\
&\hat{\bar{g}}^t_i = \sum_{t^{'}=0}^{t}\prod_{j=t^{'}+1}^{t}\roundBrack{1-\alpha_j}\alpha_{t^{'}} \tilde{\bar{g}}^{t^{'}}_i. \notag
\end{align}
Then
\begin{align}
&\norm{\hat{g}^t_i - \hat{\bar{g}}^t_i}_2 \leq \sum_{t^{'}=0}^{t}\prod_{j=t^{'}+1}^{t}\roundBrack{1-\alpha_j}\alpha_{t^{'}} e_{t^{'}} \notag\\
\leq&\sum_{t^{'}=0}^{t}\roundBrack{1-\alpha_t}^{t-t^{'}}\alpha_{t^{'}} e_{t^{'}} \notag\\
=&\sum_{t^{'}=0}^{n_t}\roundBrack{1-\alpha_t}^{t-t^{'}}\alpha_{t^{'}} e_{t^{'}} + \sum_{t^{'}=n_t+1}^{t}\roundBrack{1-\alpha_t}^{t-t^{'}}\alpha_{t^{'}} e_{t^{'}} \notag\\
\leq& e_{t_a}\frac{\roundBrack{1-\alpha_t}^{t-n_t}}{\alpha_t} + e_{t_b}\frac{\alpha_{n_t+1}}{\alpha_t},	
\end{align}
where $ e_{t^{'}} = \norm{\tilde{g}^{t^{'}}_i - \tilde{\bar{g}}^{t^{'}}_i}_2 = O\roundBrack{\abs{\hat{J}^{t^{'}}_i - J_i\roundBrack{\theta^{t^{'}}}}} $, $ n_t = \roundBrack{1-\kappa-\epsilon}t $ with $ \epsilon\in\roundBrack{0,1-\kappa} $, $ e_{t_a} = \max_{t^{'}=0,\dots,n_t} e_{t^{'}} $ and $ e_{t_b} = \max_{t^{'}=n_t+1,\dots,t} e_{t^{'}} $. From Assumption \ref{Assmp: stepsizes}-1), we have $ \lim_{t\to\infty}e_{t_a}\frac{\roundBrack{1-\alpha_t}^{t-n_t}}{\alpha_t} = 0 $ and $ \frac{\alpha_{n_t+1}}{\alpha_t}<\infty $. Then it follows from the above analysis that
\begin{equation}
\lim_{t\to\infty}\norm{\hat{g}^t_i - \hat{\bar{g}}^t_i}_2 = O\roundBrack{\lim_{t_b\to\infty}\abs{\hat{J}^{t_b}_i - J_i\roundBrack{\theta^{t_b}}}} = 0,
\end{equation}
which completes the proof of \eqref{eqn: step 2}.

In summary, it follows from the analysis in Appendix \ref{sec: step 1} and \ref{sec: step 2} that the asymptotic consistency of gradient \eqref{eqn: gradient consistency} is satisfied. Now we complete the proof of Lemma \ref{lemma: asymptotic consistency}.

\section{Proof of Theorem \ref{thm: convergence of COPO}}\label{Proof:Theorem of convergence}
1. We first prove a lemma that is crucial for the convergence proof. 
\begin{lemma}\label{lem:keylem}
	Suppose Assumptions \ref{Problem Assumptions} and \ref{Assmp: stepsizes} are satisfied, and assume $\Theta_{A}^{*}\cap\overline{\Theta}_{C}^{*}=\emptyset$,
	where $\Theta_{A}^{*}$ is the set of limiting points of Algorithm \ref{algo: SCAPO}. Denote $\left\{ \theta^{t}\right\} _{t=1}^{\infty}$ as the sequence of iterates generated by Algorithm \ref{algo: SCAPO}, we have 
	\begin{align*}
	\limsup_{t\rightarrow\infty} J\roundBrack{\theta^t}&\leq 0,\mbox{ w.p.1.}\\
	\lim_{t\rightarrow\infty}\norm{\bar{\theta}^{t}-\theta^{t}}&=0,\mbox{ w.p.1.},
	\end{align*}
	where $ J\roundBrack{\theta} $ is the maximum constraint defined in \eqref{eqn: max constraints}.	
\end{lemma}
The lemma states that when $\Theta_{A}^{*}\cap\overline{\Theta}_{C}^{*}=\emptyset$, the SCAOPO algorithm will converge to the feasible region, and the gap between $\bar{\theta}^{t}$ and $\theta^{t}$ converges to zero almost surely. Please refer to Appendix \ref{Proof: keylemma} for the proof.

2. Then we prove that under the conditions in Theorem \ref{thm: convergence of COPO}, we have $\theta^{t}\notin\overline{\Theta}_{C}^{*},\forall t$ and thus $\Theta_{A}^{*}\cap\overline{\Theta}_{C}^{*}=\emptyset$ holds true with probability 1.

When $\overline{\Theta}_{C}^{*}=\emptyset$, $\theta^{t}\notin\overline{\Theta}_{C}^{*},\forall t$ is obviously satisfied. Therefore, we focus on the non-trivial case when $\overline{\Theta}_{C}^{*}\neq\emptyset$. 
 Let
\[
\mathcal{L}\left(x\right)=\left\{ \theta:\:J\left(\theta\right)\leq x\right\} 
\]
denote a sublevel set of $J\left(\theta\right)$ at level
$x$. Let $x_{C}=\min_{\theta\in\overline{\Theta}_{C}^{*}}\:J\left(\theta\right)$.
By the definition of $\overline{\Theta}_{C}^{*}$, we must have
$x_{C}>0$. Since the initial point is feasible, i.e., $J\left(\theta^{0}\right)\leq0$, we
must have $\theta^{0}\in\mathcal{L}\left(0.5x_{C}\right)$.
Let $\Theta_{S}$ be a compact subset of $\mathcal{L}\left(0.5x_{C}\right)$
such that all the points in $\Theta_{S}$ is connected with $\theta^{0}$.
Note that by definition, $\Theta_{S}\cap\overline{\Theta}_{C}^{*}=\emptyset$.
Let $\hat{x}=0.25x_{C}$ and $\mathcal{L}\left(\hat{x},\Theta_{S}\right)=\mathcal{L}\left(\hat{x}\right)\cap\Theta_{S}$.
Since $J\left(\theta\right)$ is Lipschitz continuous following from the Assumption \ref{Problem Assumptions}-3), there
exists a constant $L>0$ such that 
\begin{equation}
\min_{\theta\in\partial\mathcal{L}\left(\hat{x},\Theta_{S}\right)}\left\Vert \theta-\theta^{0}\right\Vert \geq L\left(0.25x_{C}-J\left(\theta^{0}\right)\right)\geq0.25Lx_{C},\label{xxp-1}
\end{equation}
where $\partial\mathcal{X}$ denotes the boundary of a set $\mathcal{X}$.

By redefining the set $\mathcal{T}_{\epsilon},\mathcal{T}_{\epsilon}^{'}$
in Appendix \ref{Proof: keylemma} as $\mathcal{T}_{\epsilon}=\left\{ t:\:J\left(\theta^{t}\right)\geq\epsilon,\theta^{t}\in\Theta_{S}\right\} $,
$\mathcal{T}_{\epsilon}^{'}=\mathcal{T}_{\epsilon}\cap\left\{ t\geq t_{\epsilon}\right\} $,
and following the same analysis as in Appendix \ref{Proof: keylemma},
it can be shown that (\ref{eq:gapf}) and (\ref{eq:gapf1}) still
hold since $\Theta_{S}\cap\overline{\Theta}_{C}^{*}=\emptyset$.
Suppose we choose the initial step size $\beta_{0}<0.25Lx_{C}/\left(R_{\Theta}t_{\epsilon}\right)$,
where $R_{\Theta}\triangleq\max_{\theta, y\in\Theta}\left\Vert \theta-y\right\Vert $
is the diameter of $\Theta$. Then from \eqref{eqn: policy update}
and (\ref{xxp-1}), we must have $\theta^{t}\in\mathcal{L}\left(\hat{x},\Theta_{S}\right)$
for $t\leq t_{\epsilon}$.

From (\ref{eq:gapf}), we know that $J\left(\theta^{t}\right)$
will be decreased (almost surely) whenever $J\left(\theta^{t}\right)\geq\epsilon$,
$t\geq t_{\epsilon}$ and $\theta^{t}\in\Theta_{S}$.
Moreover, from the Lipschitz continuity of $J\left(\theta\right)$,
we have
\begin{equation}
\min_{\theta\in\mathcal{L}\left(\hat{x},\Theta_{S}\right),\theta^{'}\in\partial\Theta_{S}}\left\Vert \theta-\theta^{'}\right\Vert \geq0.25Lx_{C}.\label{xxp}
\end{equation}
Since $x_{C}>0$, we can always choose a sufficiently small $\epsilon$
such that $J\left(\theta\right)>3\epsilon,\forall\theta\in\partial\Theta_{S}$,
$\hat{x}>2\epsilon$ and $0.25Lx_{C}>2\epsilon$. From (\ref{xxp}),
once $\theta^{t}\in\mathcal{L}\left(\hat{x},\Theta_{S}\right)$
for $t\geq t_{\epsilon}$, $\theta^{t+1}$ must also belong
to $\Theta_{S}$ because $\left\Vert \theta^{t+1}-\theta^{t}\right\Vert \leq O(\beta_{t})<\epsilon$
for sufficiently large $t_{\epsilon}$, and there are two cases. 

Case 1: $J\left(\theta^{t}\right)\geq\epsilon$. In this case,
we have $J\left(\theta^{t+1}\right)<J\left(\theta^{t}\right)$
according to (\ref{eq:gapf}) and thus $\theta^{t+1}\in\mathcal{L}\left(\hat{x},\Theta_{S}\right)$
according to the definition of sublevel set, with probability 1. 

Case 2: $J\left(\theta^{t}\right)<\epsilon$. From (\ref{eq:gapf1}),
we have $J\left(\theta^{t+1}\right)<2\epsilon$ and thus $\theta^{t+1}\in\mathcal{L}\left(\hat{x},\Theta_{S}\right)$,
with probability 1. 

In any case, we have $\theta^{t+1}\in\mathcal{L}\left(\hat{x},\Theta_{S}\right)$
with probability 1. Therefore, once $\theta^{t}\in\mathcal{L}\left(\hat{x},\Theta_{S}\right)$
for $t\geq t_{\epsilon}$, it remains in $\mathcal{L}\left(\hat{x},\Theta_{S}\right)$
with probability 1. Together with the fact that $\theta^{t}\in\mathcal{L}\left(\hat{x},\Theta_{S}\right),\forall t\leq t_{\epsilon}$,
we conclude that $\theta^{t}\in\mathcal{L}\left(\hat{x},\Theta_{S}\right)\subset\Theta_{S},\forall t$
with probability 1. Since $\Theta_{S}\cap\overline{\Theta}_{C}^{*}=\emptyset$,
we have $\theta^{t}\notin\overline{\Theta}_{C}^{*},\forall t$
with probability 1.

3. Finally, we prove Theorem \ref{thm: convergence of COPO}.

Let $\left\{ \theta^{t_{j}}\right\} _{j=1}^{\infty}$ denote
any subsequence converging to a limiting point $\theta^{*}$
that satisfies the Slater condition. Since $\theta^{t}\notin\overline{\Theta}_{C}^{*},\forall t$
and $\Theta_{A}^{*}\cap\overline{\Theta}_{C}^{*}=\emptyset$
w.p.1., it follows from Lemma \ref{lem:keylem} that 
\begin{equation}
\lim_{j\rightarrow\infty}\left\Vert \bar{\theta}^{t_{j}}-\theta^{t_{j}}\right\Vert =0,\text{ w.p.1.},\label{eq:lem2rst}
\end{equation}
and
\begin{align}
\bar{\theta}^{t_{j}}=\underset{\theta\in\Theta}{\text{argmin}}\: & \bar{J}_{0}^{t_{j}}\left(\theta\right)\label{eq:Pixbart-1}\\
\st\: & \bar{J}_{i}^{t_{j}}\left(\theta\right)\leq x^{t_{j}},i=1,....,m,\nonumber 
\end{align}
where 
\begin{equation}
\lim_{j\rightarrow\infty}x^{t_{j}}=0,\text{ w.p.1.}\label{eq:alpha0}
\end{equation}
Using (\ref{eq:lem2rst}), (\ref{eq:Pixbart-1}), (\ref{eq:alpha0}), Lemma \ref{lemma: asymptotic consistency}
and the strong convexity of $\bar{J}_{i}^{t}\left(\theta\right),\hat{J}_{i}\left(\theta\right),\forall i$,
we have
\begin{align}\label{eq:Piterthead}
\theta^{*}=\underset{\theta\in\Theta}{\text{argmin}}\: & \hat{J}_{0}\left(\theta\right)\nonumber\\
s.t.\: & \hat{J}_{i}\left(\theta\right)\leq 0,i=1,....,m. 
\end{align}
Then it follows from Lemma \ref{lemma: asymptotic consistency}  and (\ref{eq:Piterthead}) that $\theta^{*}$ satisfies the KKT condition of Problem \eqref{original_problem}. This completes the proof.

\section{Proof of Lemma \ref{lem:keylem}}\label{Proof: keylemma}

1. We first prove $\limsup_{t\rightarrow\infty}J\left(\theta^{t}\right)\leq0$ w.p.1. 

Let $\mathcal{T}_{\epsilon}=\left\{ t:\:J\left(\theta^{t}\right)\geq\epsilon\right\} $
for any $\epsilon>0$. We show that $\mathcal{T}_{\epsilon}$ is a
finite set by contradiction. 

Suppose $\mathcal{T}_{\epsilon}$ is infinite. We first show that
$\liminf_{t\in\mathcal{T}_{\epsilon},t\rightarrow\infty}\left\Vert \bar{\theta}^{t}-\theta^{t}\right\Vert >0$
by contradiction. Suppose $\liminf_{t\in\mathcal{T}_{\epsilon},t\rightarrow\infty}\left\Vert \bar{\theta}^{t}-\theta^{t}\right\Vert =0$.
Then there exists a subsequence $t^{j}\in\mathcal{T}_{\epsilon}$
such that $\lim_{j\rightarrow\infty}\left\Vert \bar{\theta}^{t_{j}}-\theta^{t_{j}}\right\Vert =0$.
Let $\theta^{\circ}$ denote a limiting point of the subsequence
$\left\{ \theta^{t_{j}}\right\} $, there are two cases according to the update rule of Algorithm \ref{algo: SCAPO}.
If $\theta^{\circ}$ is the optimal solution of the objective update, it is obviously $J\left(\theta^{\circ}\right)\leq0$ following from Lemma \ref{lemma: asymptotic consistency}. If $\theta^{\circ}$ is the optimal solution of the feasible update, it can be verified that $\theta^{\circ}$  satisfies the KKT condition of Problem \eqref{eqn: max constraints} following from Lemma \ref{lemma: asymptotic consistency}. From the condition $\Theta_{A}^{*}\cap\overline{\Theta}_{C}^{*}=\emptyset$, we have $J\left(\theta^{\circ}\right)\leq0$. The above two cases both contradicts the definition of $\mathcal{T}_{\epsilon}$.




Therefore, $\liminf_{t\in\mathcal{T}_{\epsilon},t\rightarrow\infty}\left\Vert \bar{\theta}^{t}-\theta^{t}\right\Vert >0$,
i.e., there exists a sufficiently large $t_{\epsilon}$ such that
\begin{equation}
\left\Vert \bar{\theta}^{t}-\theta^{t}\right\Vert \geq\epsilon^{'},\forall t\in\mathcal{T}_{\epsilon}^{'}\label{eq:gapxt}
\end{equation}
where $\epsilon^{'}>0$ is some constant and $\mathcal{T}_{\epsilon}^{'}=\mathcal{T}_{\epsilon}\cap\left\{ t\geq t_{\epsilon}\right\} $. 

Define function $\bar{J}^{t}\left(\theta\right)\triangleq\max_{i\in\left\{ 1,...,m\right\} }\bar{J}_{i}^{t}\left(\theta\right)$. $\bar{J}_{i}^{t}\left(\theta^{t}\right)$
is strongly convex, and thus
\begin{equation}
\nabla^{T}\bar{J}_{i}^{t}\left(\theta^{t}\right)d^{t}\leq-\eta\left\Vert d^{t}\right\Vert ^{2}+\bar{J}_{i}^{t}\left(\bar{\theta}^{t}\right)-\bar{J}_{i}^{t}\left(\theta^{t}\right),\label{eq:ftdbound}
\end{equation}
where $d^{t}=\bar{\theta}^{t}-\theta^{t}$,
and $\eta>0$ is some constant. From Assumption \ref{Problem Assumptions}-3),
the gradient of $J_{i}\left(\theta\right)$ is Lipschitz continuous,
and thus there exists $L_{J}>0$ such that
\begin{align}
J_{i}\left(\theta^{t+1}\right) & \leq J_{i}\left(\theta^{t}\right)+\beta_{t}\nabla^{T}J_{i}\left(\theta^{t}\right)d^{t}+L_{J}\left(\beta_{t}\right)^{2}\left\Vert d^{t}\right\Vert^{2}_2\nonumber \\
& =J\left(\theta^{t}\right)+L_{J}\left(\beta_{t}\right)^{2}\left\Vert d^{t}\right\Vert^{2}_2 + J_{i}\left(\theta^{t}\right)-J\left(\theta^{t}\right)\nonumber \\
& +\beta_{t}\left(\nabla^{T}\bar{J}_{i}^{t}\left(\theta^{t}\right)+\nabla^{T}J_{i}\left(\theta^{t}\right)-\nabla^{T}\bar{J}_{i}^{t}\left(\theta^{t}\right)\right) d^{t}\nonumber \\
& \overset{\textrm{a}}{\leq}J\left(\theta^{t}\right)+J_{i}\left(\theta^{t}\right)-J\left(\theta^{t}\right)-\eta\beta_{t}\left\Vert d^{t}\right\Vert^{2}_2\nonumber \\
& +\beta_{t}\left(\bar{J}_{i}^{t}\left(\bar{\theta}^{t}\right)-\bar{J}_{i}^{t}\left(\theta^{t}\right)\right)+o\left(\beta_{t}\right)\nonumber \\
& \leq J\left(\theta^{t}\right)-\eta\beta_{t}\left\Vert d^{t}\right\Vert^{2}_2 + o\left(\beta_{t}\right),\forall i=1,...,m\label{eq:fxdecre}
\end{align}
where $o\left(\beta_t\right)$ means that $\lim_{t\rightarrow\infty}o\left(\beta_t\right)/\beta_t=0$.
In (\ref{eq:fxdecre}-a), we use (\ref{eq:ftdbound}) and $\lim_{t\rightarrow\infty}\left\Vert \nabla J_{i}\left(\theta^{t}\right)-\nabla\bar{J}_{i}^{t}\left(\theta^{t}\right)\right\Vert =0$,
and the last inequality follows from $J_{i}\left(\theta^{t}\right)\leq J\left(\theta^{t}\right)$,
$\liminf_{t\rightarrow\infty}J\left(\theta^{t}\right)-\bar{J}_{i}^{t}\left(\bar{\theta}^{t}\right)\geq0$,
and $\lim_{t\rightarrow\infty}\left\Vert J_{i}\left(\theta^{t}\right)-\bar{J}_{i}^{t}\left(\theta^{t}\right)\right\Vert =0$.
Since (\ref{eq:fxdecre}) holds for all $i=1,...,m$, by choosing
a sufficiently large $t_{\epsilon}$, we have 
\begin{align}
J\left(\theta^{t+1}\right)-J\left(\theta^{t}\right) & \leq-\beta_{t}\overline{\eta}\left\Vert d^{t}\right\Vert^{2}_2\nonumber \\
& \leq-\beta_t\overline{\eta}\epsilon^{'},\forall t\in\mathcal{T}_{\epsilon}^{'}.\label{eq:gapf}
\end{align}
for some $\overline{\eta}>0$. Moreover,
$J\left(\theta\right)$ is Lipschitz continuous, and thus
\begin{equation}
\left|J\left(\theta^{t+1}\right)-J\left(\theta^{t}\right)\right|\leq O(\left\Vert \theta^{t+1}-\theta^{t}\right\Vert_2 )\leq O(\beta_t)<\epsilon,\label{eq:gapf1}
\end{equation}
$\forall t\geq t_{\epsilon}$, for sufficiently large $t_{\epsilon}$,
where the last inequality follows from $\beta_t\rightarrow0$ as
$t\rightarrow\infty$. From (\ref{eq:gapf}), we know that $J\left(\theta^{t}\right)$
will be decreased (almost surely) whenever $J\left(\theta^{t}\right)\geq\epsilon$
and $t\geq t_{\epsilon}$. Therefore, it follows from (\ref{eq:gapf})
and (\ref{eq:gapf1}) that 
\begin{equation}
J\left(\theta^{t}\right)\leq2\epsilon,\forall t\geq t_{\epsilon}.\label{eq:fxtbound}
\end{equation}
Since (\ref{eq:fxtbound}) is true for any $\epsilon>0$, it follows
that $\limsup_{t\rightarrow\infty}J\left(\theta^{t}\right)\leq0$.

2. Then we prove that $\lim_{t\rightarrow\infty}\left\Vert \bar{\theta}^{t}-\theta^{t}\right\Vert =0,$
w.p.1. 

According to the strong convexity of $\bar{J}_{i}^{t}\left(\theta\right)$, the boundness of $J_{i}\left(\theta\right)$, the Lipschitz continuity of $\nabla J_{i}\left(\theta\right)$ for $ i\in\curlyBrack{0,\dots,m} $ and $ \sum_{t} \beta_t = \infty $, we can prove $\lim_{t\rightarrow\infty}\left\Vert \bar{\theta}^{t}-\theta^{t}\right\Vert =0,$ w.p.1. following the similar analysis in Appendix D-2 of \cite{2019CSSCA}. We omit the proof due to the space limit.

\ifCLASSOPTIONcaptionsoff
  \newpage
\fi

\bibliographystyle{IEEEtran}
\bibliography{IEEEabrv,References}
\end{document}